\documentclass[letterpaper]{article} 
\usepackage{aaai24}  
\usepackage{times}  
\usepackage{helvet}  
\usepackage{courier}  
\usepackage[hyphens]{url}  
\usepackage{graphicx} 
\usepackage{natbib}  
\usepackage{caption} 
\usepackage{appendix}
\usepackage{graphicx}
\usepackage{float}
\usepackage{amsmath}
\usepackage{array}
\usepackage{subcaption}
\usepackage[table]{xcolor} 

\nocopyright 
\usepackage{booktabs}
\usepackage{subcaption}

\title{Can LLMs plan paths in the real world?}
\author {
    Wanyi Chen\textsuperscript{\rm 1},
    Meng-Wen Su \textsuperscript{\rm 2}, 
    Nafisa Mehjabin \textsuperscript{\rm 2}, 
    Mary L. Cummings\textsuperscript{\rm 2}
}
\affiliations {
    \textsuperscript{\rm 1}Duke University, 
    \textsuperscript{\rm 2}George Mason University
}

\begin{document}
\maketitle

\begin{abstract}
    As large language models (LLMs) increasingly integrate into vehicle navigation systems, understanding their path-planning capability is crucial. We tested three LLMs through six real-world path-planning scenarios in various settings and with various difficulties. Our experiments showed that all LLMs made numerous errors in all scenarios, revealing that they are unreliable path planners. We suggest that future work focus on implementing mechanisms for reality checks, enhancing model transparency, and developing smaller models.
\end{abstract}

\section{Introduction}
In early 2024, Volkswagen premiered the first vehicle with ChatGPT integrated into its voice assistant \cite{VW}. Volkswagen claimed that its ChatGPT-enabled voice assistant could be used to control the infotainment, navigation, and air conditioning, or to answer general knowledge questions \cite{VW}. Soon after, Mercedes-Benz, BMW, and Hyundai all announced plans to install voice assistants powered by LLMs in vehicles \cite{benz}. If LLM-powered voice assistants are to control vehicle navigation, it becomes important to assess whether the LLMs can plan paths in the real world.


To investigate this, we tested three LLMs of various sizes: GPT-4, Gemini, and Mistral. We created six path-planning scenarios, including typical vehicle-based paths but also visual landmark paths for pedestrians. Each scenario included finding a path from a local university campus starting point to a destination of interest, sometimes with time constraints. Each LLM was tasked to generate turn-by-turn instructions or instructions based on visual landmarks. We analyze their results and discuss whether LLMs can be reliable path planners in the real world.

\section{Background and related work}

Recently, there has been a surge of interest in exploring whether LLMs can plan paths. Effective and efficient path planning is a paramount challenges in robotics \cite{latif20243p}. Traditional path-planning algorithms suffer from memory and computational inefficiencies as the map size increases \cite{meng2024llm}. Therefore, researchers search for alternatives, and LLMs have captured their attention as an emergent technology. While LLMs have shown promise in many word processing tasks, it is unclear as to whether they can also solve path-planning problems.

Researchers have proposed various methods of utilizing LLMs in path planning. For example, \citet{latif20243p} translated robot perception to natural language so that an LLM could plan a path, and then translated this natural language to robot control. Others seek to use LLMs in combination with traditional path-planning algorithms. For instance, \citet{meng2024llm} proposed LLM-A*. They used LLMs to generate waypoints in between a start state and a goal state, and then used A*, a traditional path-planning algorithm, to plan paths connecting those waypoints \cite{meng2024llm}.

In addition, researchers have explored how LLMs can be used in vision-and-language navigation (VLN). In robotics, VLN involves giving robots or agents verbal instructions on how to navigate using visual cues and landmarks \cite{schumann2024velma}. Challenges of VLN include visual and natural language understanding, as well as spatial and temporal reasoning  \cite{schumann2024velma}. Attempting to overcome these challenges, \citet{schumann2024velma} proposed VELMA, an embodied LLM agent that uses verbalization of the trajectory and visual environment observations as prompts for the next action. Similarly, \citet{zhou2024navgpt} introduced NavGPT, which is a purely LLM-based navigation agent that follows VLN instructions.


Researchers have yet to reach an agreement as to whether LLMs are good path planners. Some are very optimistic about LLMs’ path-planning ability. For instance, \citet{latif20243p} claimed that GPT-3.5-turbo outperforms state-of-the-art path planners like Rapidly Exploring Random Tree and A*. \citet{zhou2024navgpt} claimed that NavGPT can perform high-level planning for navigation and has reasoning capability in complex scenes.

However, other researchers recognize LLMs’ limitations in path planning. For example, \citet{valmeekam2023planning} found that an LLM’s ability to autonomously generate executable plans is lacking, averaging only about 3\% success rate. \citet{aghzal2023can} also found that LLMs struggle to generalize to larger environments or those with more obstacles. Furthermore, \citet{kambhampati2024can} argued that LLMs rely a lot on fine-tuning and/or human-in-the-loop prompting to achieve good results.

While these studies led to valuable insights into LLMs' path-planning abilities, they were conducted in simulated environments, which are often oversimplifications of complicated real-world scenarios. Our research contributes to the ongoing debate by testing LLMs' path-planning abilities in the real world. The next section introduces our experimental scenarios, the LLMs tested, and the LLM prompt design.

\section{Method}

\begin{table*}[h!]
\centering
\resizebox{\textwidth}{!}{%
\begin{tabular}{|l|l|c|l|p{6cm}|}
\hline
\textbf{Model} & \textbf{Developer} & \textbf{Parameters (approx.)} & \textbf{Accessibility} & \textbf{Key Features} \\ \hline
GPT-4  & OpenAI             & 1.8 trillion                  & Not Open-source, API access only        & Optimized for general-purpose tasks including reasoning and natural language generation  \\ \hline
Gemini         & Google DeepMind    & 70 billion                    & Not Open-source, API access only        & Focuses on contextual understanding, excels in instruction-following and interpretation \\ \hline
Mistral 7B     & Mistral AI         & 7.3 billion                   &  Open-source model, Downloadable           & Balances efficiency with performance in interpretive and generative tasks             \\ \hline
\end{tabular}%
}
\caption{Comparison of Large Language Models Tested}
\label{table:llm_comparison}
\end{table*}

We structured our experiments into two categories: Turn-by-Turn (TbT) navigation and VLN. We designed each category to test navigation across different environments and difficulty levels, ensuring the LLMs were challenged in diverse real-world contexts. 

\subsubsection{TbT navigation:} Models were instructed to provide step-by-step directions, similar to GPS navigation. We designed three TbT scenarios across various environments:

\begin{itemize}
    \item \textbf{Urban Scenario:} The urban scenario involved guiding users from George Mason University’s Shenandoah Parking Deck to the Washington Nationals Stadium. This route was set in a dense city environment, including frequent turns and high traffic density.

    \item \textbf{Suburban Scenario:}  This route from George Mason University’s Shenandoah Parking Deck to the Hogsback Mountain Paintball Center required navigation through mixed-density areas.

    \item \textbf{Rural Scenario:} In a sparsely populated rural setting, models were asked to provide directions from George Mason University’s Shenandoah Parking Deck to the Shot Tower State Park. This was considered a very challenging task since the Shot Tower State Park is located within another larger state park.
\end{itemize}

\subsubsection{VLN:} Models were tasked to generate directions using visual landmark cues instead of road signs. We designed three scenarios with varying difficulties:


\begin{itemize}
    \item \textbf{Easy Scenario:} The easy visual landmark scenario included guiding users from the Nguyen Engineering Building to the Johnson Center at George Mason University using easily recognizable landmarks like statues, building names, and common student paths.
\end{itemize}

\begin{itemize}
    \item \textbf{Medium Scenario:}  The medium scenario involved a route from the Nguyen Engineering Building to the INTO Mason Center for international students, with moderate complexity. This scenario required models to use intermediate landmarks, such as unique architectural features or pathway markers, to provide directions for users unfamiliar with road signs.
\end{itemize}

\begin{itemize}
    \item \textbf{Hard Scenario:} The hard VLN scenario required models to navigate from the Nguyen Engineering Building to the Osher Lifelong Learning Institute (OLLI). This route involved more challenging navigation across  and off the university campus, with fewer clear landmarks. It tested each model’s ability to identify subtle features or less commonly recognized landmarks for effective guidance.
\end{itemize}

To test whether model size impacts LLM's path-planning abilities, we selected three LLMs of varying scales. Table \ref{table:llm_comparison} shows a comparative summary of the LLMs. Each LLM was tested across all six scenarios, allowing for a detailed comparison of their performances. 

\subsection{Prompt Design} For the TbT scenarios, we designed prompts to test whether the LLMs could plan paths that meet user-specified time constraints. Each prompt specified a start and end point, providing context for the model to generate timely, navigational instructions based on user needs. For example, a user may want to arrive at a stadium two hours before a game. If the LLM failed to provide a detailed response to the initial query, we asked follow-up questions.  Approximately 77.8\% of TbT scenarios required at least one follow-up question. Below is an example of the prompt design for the urban TbT scenario. Additional prompts and their corresponding follow-up questions are provided in Appendix A. \\

\noindent \textbf{Initial Prompt-} \textit{``Plan a trip using turn-by-turn instructions from George Mason University Shenandoah Parking Deck to Garage C of the Nationals Stadium to arrive 2 hours before the first-weekend game in August 2024."} 

\noindent \textbf{GPT-4 Initial Response-}
\textit{“To provide a detailed trip plan, I need the exact date and time of the game you plan to attend. If you don't have the exact date yet, please provide an estimated time for planning purposes (e.g., an afternoon or evening game). This will help ensure you arrive 2 hours before the game.”} \\

\noindent \textbf{Follow-Up Question 1-} \textit{"What is the first weekend in August 2024?"}

\noindent \textbf{GPT-4 Response-}\textit{“The first weekend in August 2024 is August 3rd and 4th. If you are planning to attend a game on either of these dates, please specify the time of the game so I can plan your trip accordingly.”} \\

\noindent \textbf{Follow-Up Question 2-} \textit{"Now can you find if the game is on August 3rd or 4th and what time should I leave from George Mason University Shenandoah Parking Deck to arrive at  Garage C of the Nationals Stadium two hours early?"}

\noindent \textbf{GPT-4 Final Response-} \textit{“The Washington Nationals have games scheduled on the first weekend of August 2024 at Nationals Park. Here are the game details:  \newline
Saturday, August 3rd, 2024: The game starts at 4:05 PM.  \newline
Sunday, August 4th, 2024: The game starts at 1:35 PM. \\
Given these timings, to arrive 2 hours before the game, you should plan to reach Nationals Park by 2:05 PM on Saturday and by 11:35 AM on Sunday. Below are the turn-by-turn directions for your trip from George Mason University Shenandoah Parking Deck to Garage C at Nationals Park.”} \\

\noindent For VLN tasks, prompts were designed to evaluate the LLMs’ ability to provide clear, visual-based directions, appropriate for users who rely on physical landmarks rather than road signs. Each prompt specified a start and end point, asking the model to focus on notable campus landmarks to guide the user along the route. For instance, an easy-level prompt instructed the LLM to navigate from the Nguyen Engineering Building to the Johnson Center, using visible cues such as “the George Mason statue” and “Mason Pond.” In most cases, the LLM successfully generated detailed, step-by-step directions. However, for a few instances with Gemini on medium-level prompts, follow-up questions were used to encourage the model to generate richer, landmark-based guidance instead of relying on map links. Below is an example of the prompt design for the medium scenario. All prompts and responses for the VLN scenarios are included in the appendix. \\

\noindent \textbf{Initial Prompt:} \textit{Plan a trip using visual landmarks for people who cannot read road sign instructions from George Mason University Nguyen Engineering Building to INTO Mason Center.}

\noindent \textbf{Gemini Initial Response:} \textit{ Provided link of Google Map} \\

\noindent \textbf{Follow-up Question 1:} \textit{Plan a step-by-step visual landmark instruction for people who cannot read road sign instructions from George Mason University Nguyen Engineering Building to INTO Mason Center.}

\noindent \textbf{Gemini Final Response:}

\begin{itemize}
    \item \textbf{Starting Point: Nguyen Engineering Building}
    \begin{itemize}
        \item \textbf{Exit the Nguyen Engineering Building:} Walk out the main entrance facing Patriot Circle.
        \item \textbf{Turn right and walk on the Patriot Cir pathway:} You will see a large circular road with a pond in the center.
    \end{itemize}

    \item \textbf{Navigating to Mason Pond Drive}
    \begin{itemize}
        \item \textbf{Walk along Patriot Circle:} Continue walking along the sidewalk on your right, keeping the pond on your left.
        \item \textbf{Look for the Bridge:} You'll encounter a bridge crossing over a small stream.
        \item \textbf{Turn Right:} After crossing the bridge, immediately turn right onto a road that runs parallel to the pond. This is Mason Pond Drive.
    \end{itemize}

    \item \textbf{Reaching the INTO Mason Center}
    \begin{itemize}
        \item \textbf{Follow the Road:} Continue walking straight on Mason Pond Drive.
        \item \textbf{Identify the INTO Mason Center:} You'll see a modern, multi-story building on your left with the words ``INTO Mason Center" displayed prominently.
    \end{itemize}

    \item You have arrived at the INTO Mason Center.
\end{itemize}


\section{Results}
To evaluate the performance of each LLM in Table 1, we analyzed the LLM-generated responses compared to a ground-truth route from Waze, a navigation app that is primarily used on mobile devices.

\subsection{Types of Errors}
We classified LLM errors into major and minor errors. Major errors are those that could mislead users or direct them to a different destination, making it difficult for them to figure out how to reach their intended location. In contrast, minor errors are instances of misinformation or misdirection in a smaller area, where users can likely find their way back on track by recognizing signs or asking for directions. These are discussed further below:

\subsubsection{Major errors for TbT navigation:}

\begin{figure} 
    \centering
    \begin{subfigure}[b]{0.31\textwidth}         \includegraphics[width=\textwidth]{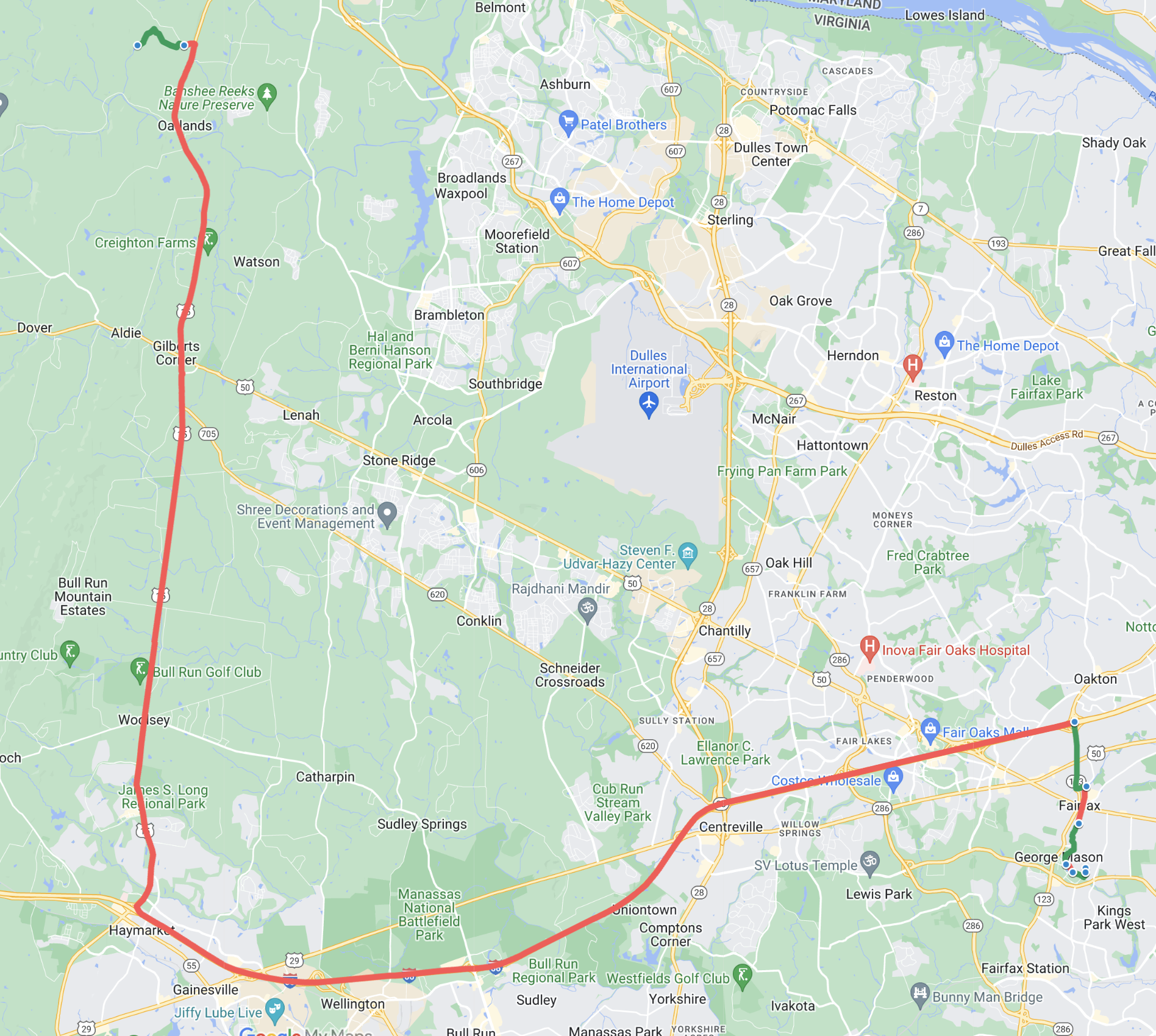}
        \caption{GPT's path}
        \label{fig:gpt_suburb}
    \end{subfigure}

    \begin{subfigure}[b]{0.31\textwidth}        \includegraphics[width=\textwidth]{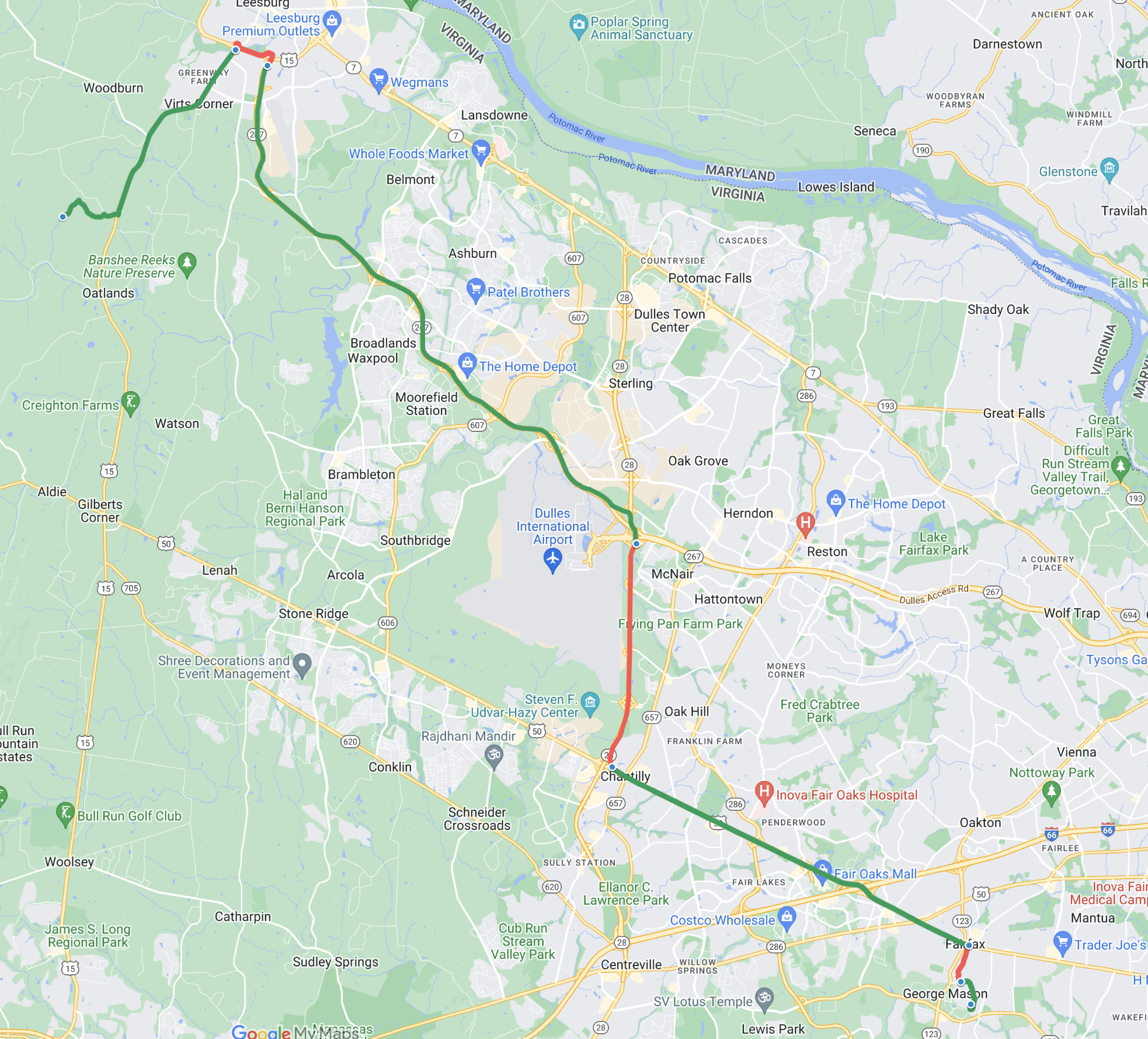}
        \caption{Gemini's path}
        \label{fig:gemini_suburb}
    \end{subfigure}

    \begin{subfigure}[b]{0.31\textwidth}        \includegraphics[width=\textwidth]{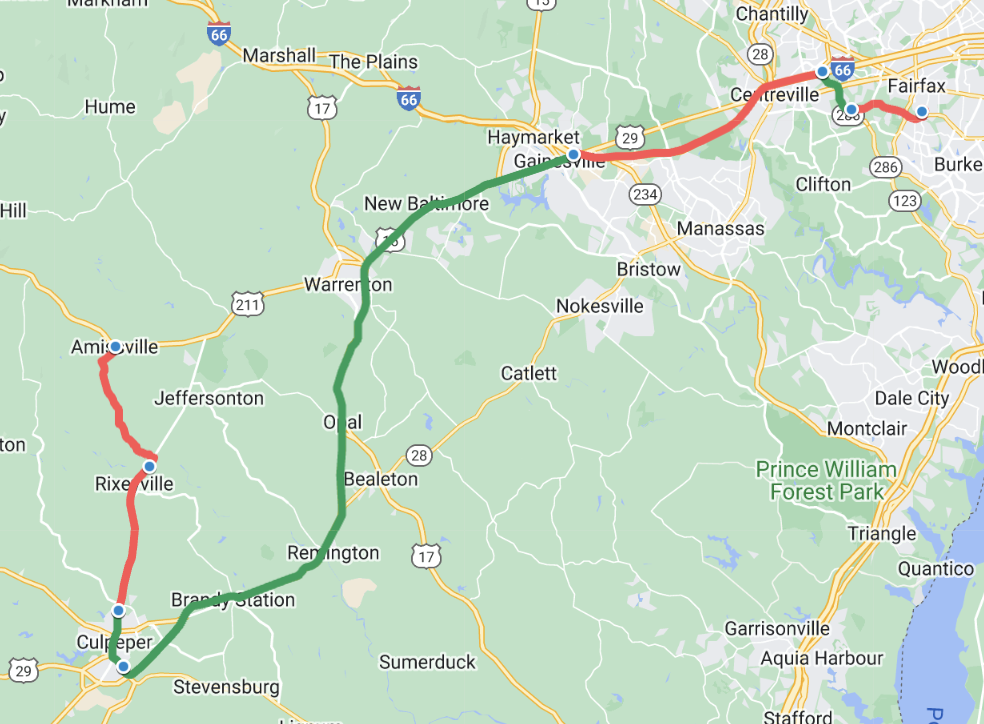}
        \caption{Mistral's path}
        \label{fig:mistral_suburb}
    \end{subfigure}

    \begin{subfigure}[b]{0.31\textwidth}        \includegraphics[width=\textwidth]{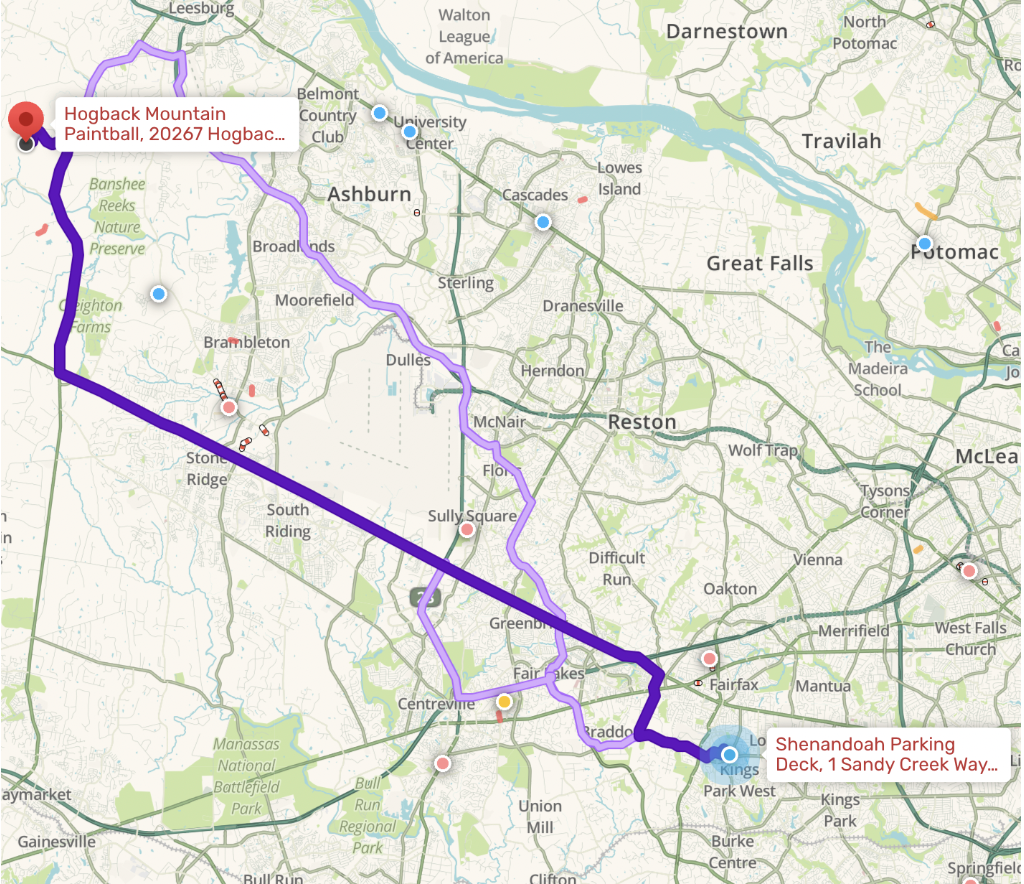}
        \caption{Waze's path (ground truth)}
        \label{fig:waze_suburb}
    \end{subfigure}

    \caption{Visualization of paths for the suburban TbT scenario. LLM-generated paths are marked in red. Discontinuities are marked in green.}
    \label{fig:suburb_tbt}
\end{figure}

\begin{enumerate}
    \item \textit{Discontinuities} - LLMs often assume that the routes they generate are continuous. However, there are often significant gaps or missing segments in these routes. In such cases, we identify the shortest path to fill these substantial gaps or missing sections. For example, in a suburban scenario, the model named Gemini assumed that University Drive connected directly to VA-28, but there is a notable gap between these two roads. After determining the shortest path that connects University Drive to VA-28, we discovered an 8.8-mile discontinuity between the routes. We have classified these and other similar route breaks as discontinuities, which are illustrated in Figure \ref{fig:suburb_tbt}a-c. The routes generated by the LLMs are marked in red, whereas the discontinuities are marked in green. Waze's route is shown in Figure \ref{fig:waze_suburb} as a comparison. In Figure \ref{fig:mistral_suburb}, we observe a major discontinuity where the model fails to reach the correct destination. The actual destination, ``Hogsback Mountain Paintball Center," is located at 20267 Hogback Mountain Rd, Leesburg, VA 20175, approximately 50.3 miles away. However, the model incorrectly generated `15540 Hogsback Rd, Amissville, VA 20106' as the destination. This suggests that the LLM mistakenly matched the location to a similar name or address, although no such address exists in Amissville, VA. The error likely arises from the model's tendency to generate plausible-sounding but incorrect details when faced with ambiguous or incomplete location information, even when the correct destination was clearly specified.
    \item \textit{Turns in Opposite Direction} - When merging onto highways or interstates, the LLM gave the opposite direction. For example, when the correct route should be I-95 North, but the LLM provided I-95 South instead.
    \item \textit{Wrong/missed Exit }- The LLM provided the wrong exit or missed an exit on the highway or interstate.
\end{enumerate}
\subsubsection{Major errors for VLN:}
\begin{enumerate}
    \item \textit{Discontinuties} - LLMs assumed that the given routes were continuous. However, there were significant gaps and missing routes in between the start end endp oints. Since the VLN study was conducted on the George Mason University campus, a discontinuity involving three or more buildings was considered a major error.
    \item \textit{Passing / Didn’t reach destination} -  The LLM never successfully guided users to the designated end points.
    \item \textit{Landmark didn’t exist / Wrong visual description} - The landmark mentioned either does not exist or does not match the actual description. To ensure the accuracy of the landmarks suggested by the LLMs, we conducted in-person validations by visiting the campus and walking the routes generated by the LLMs. For instance, if an LLM described that a building has a glass facade, we verified whether the building indeed has a glass facade. 
\end{enumerate}

\begin{figure} 
    \centering
    \begin{subfigure}[b]{0.37\textwidth}         \includegraphics[width=\textwidth]{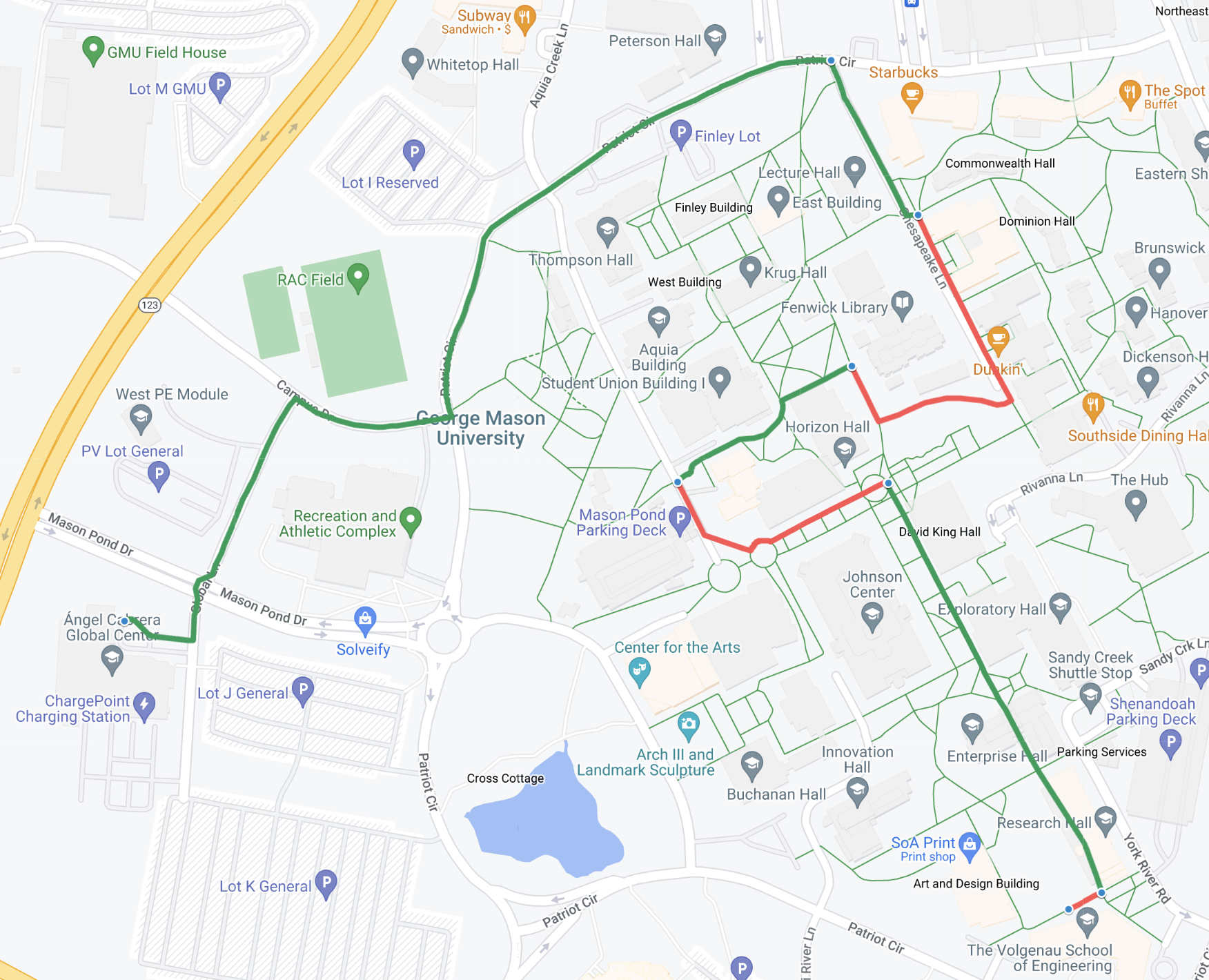}
        \caption{GPT's path}
        \label{fig:gpt_vl}
    \end{subfigure}
    \hfill
    \begin{subfigure}[b]{0.37\textwidth}        \includegraphics[width=\textwidth]{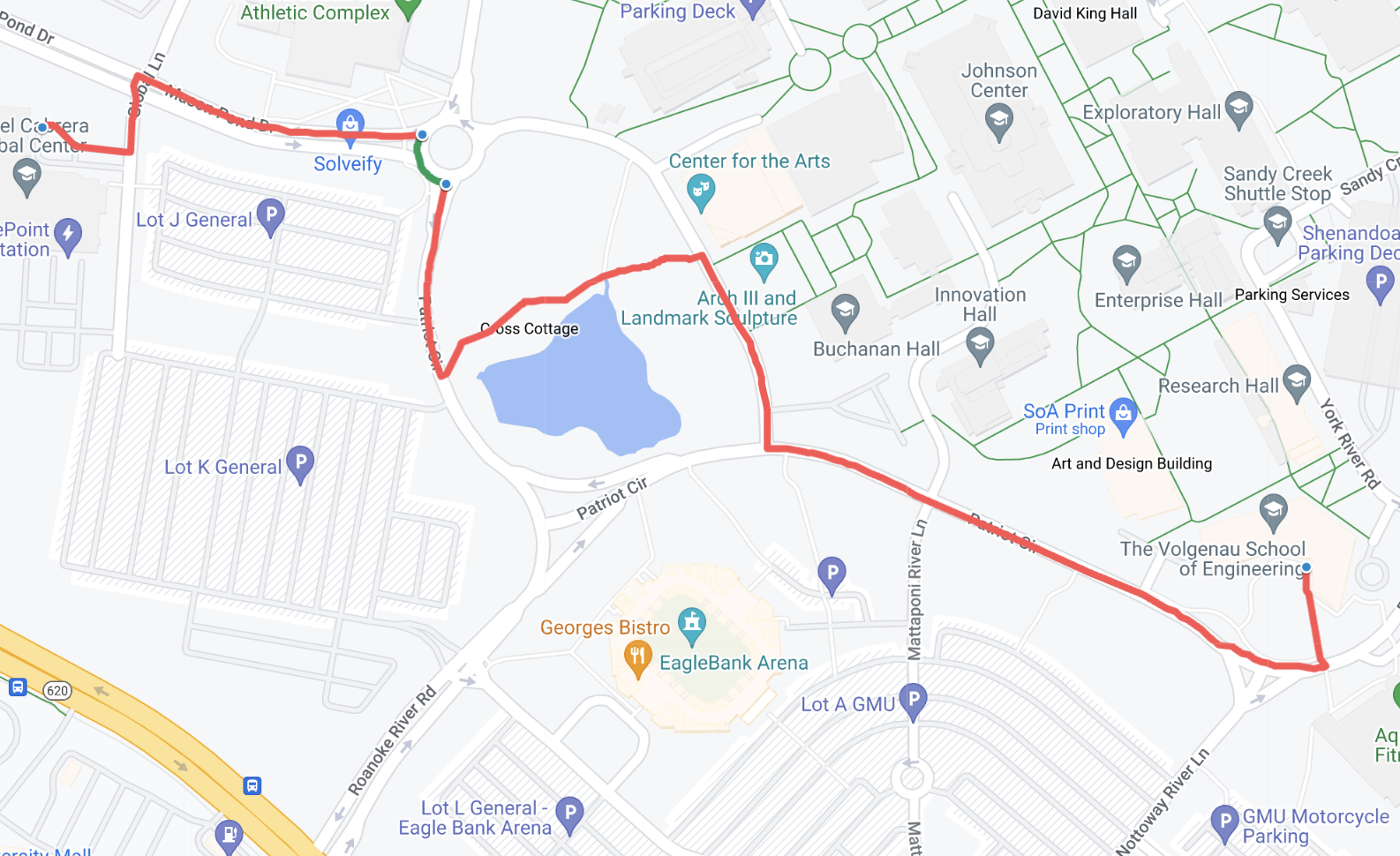}
        \caption{Gemini's path}
        \label{fig:gemini_vl}
    \end{subfigure}
    \hfill
    \begin{subfigure}[b]{0.37\textwidth}        \includegraphics[width=\textwidth]{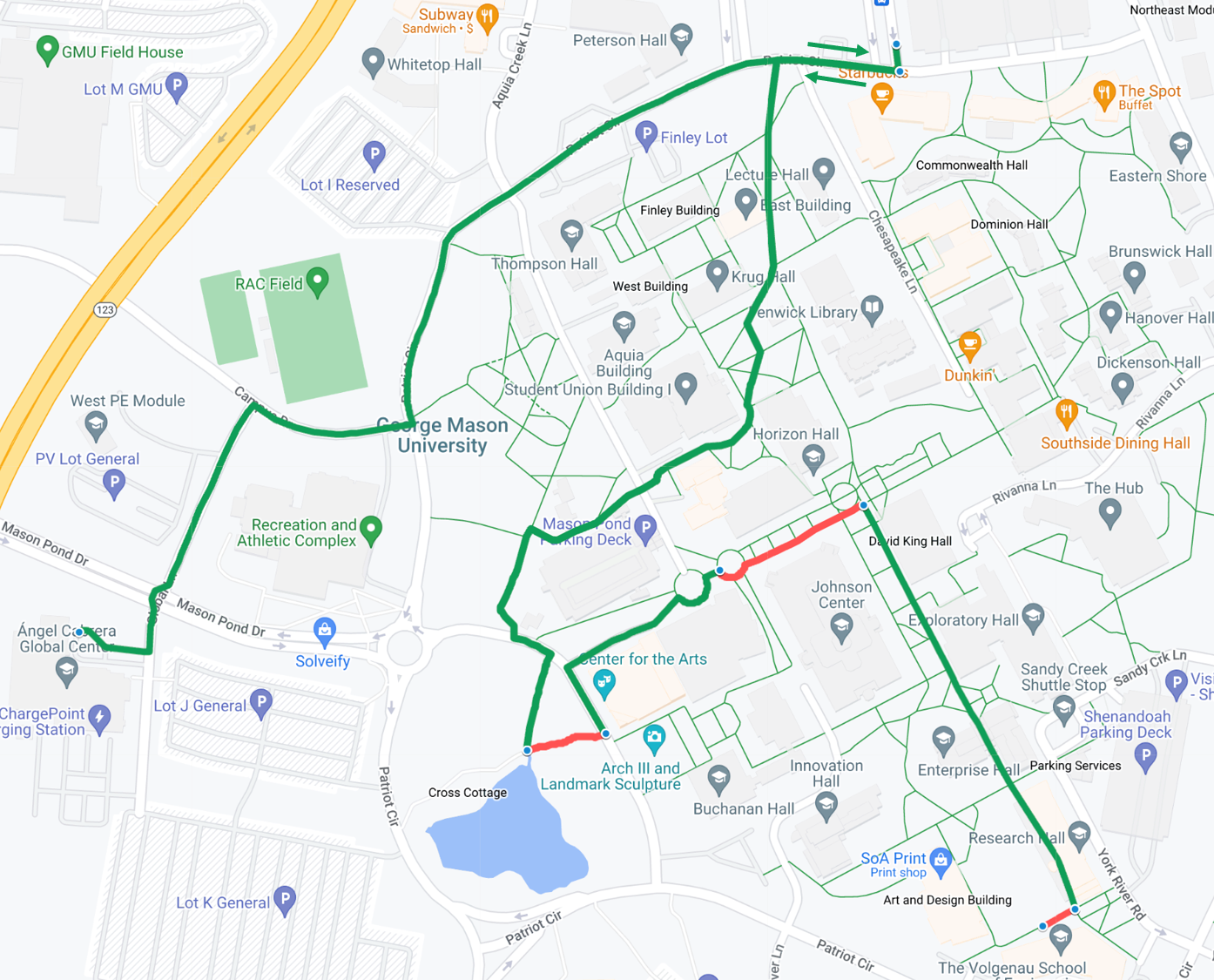}
        \caption{Mistral's path}
        \label{fig:mistral_vl}
    \end{subfigure}
    \hfill
    \begin{subfigure}[b]{0.37\textwidth}        \includegraphics[width=\textwidth]{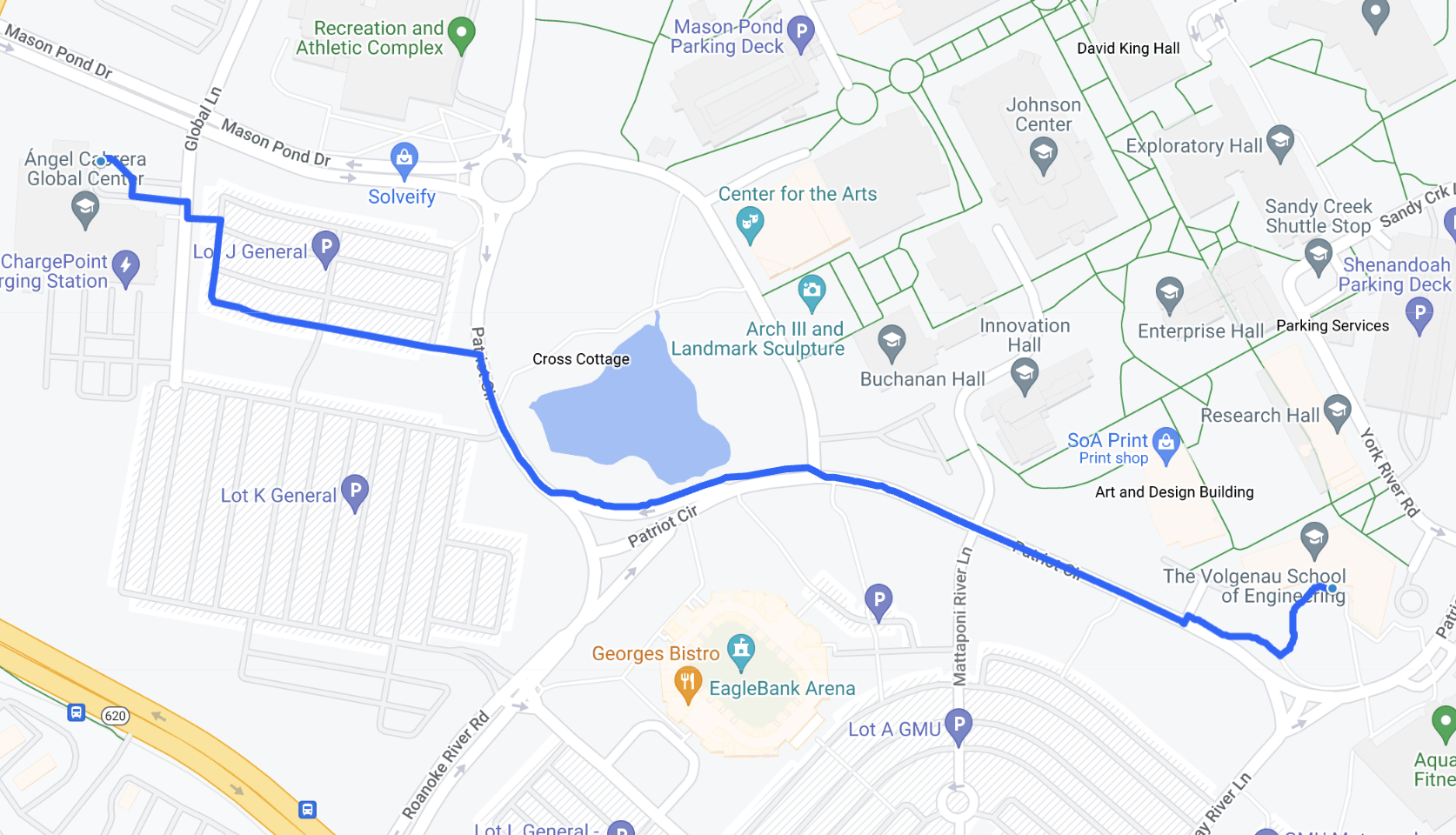}
        \caption{Typical student path}
        \label{fig:student_vl}
    \end{subfigure}

    \caption{Visualization of paths for the medium VLN scenario. LLM-generated paths are marked in red. Discontinuities are marked in green.}
\end{figure}

All other errors are classified as minor errors. For instance, in the Gemini suburban TbT scenario, there is a discontinuity between the Shenandoah Parking Deck and the next street, University Drive. This is a minor issue because it is in the local area, spanning only 0.5 miles. Drivers would likely be able to detect this problem based on their familiarity with the city or by asking for directions. Additionally, if the LLM response included the correct next street but indicated the wrong turn, we classified that as a minor error. For VLN errors, if the LLM-recommended path was off by 1 to 2 buildings, we considered that a minor error.

\subsection{Total Discontinuity and Turns}
In addition to tracking the total number of errors, we also collected data on the total discontinuous miles in the LLM routes for the TbT scenarios. We recorded the number of LLM turns to the designated locations. Furthermore, we categorized the planned routes into different levels of driver knowledge required to successfully reach the destination. This was based on the size of the discontinuity and whether it occurred in a complex area or a local area where drivers could easily navigate around:
\begin{enumerate}
    \item \textit{Beginner} - Someone who could not navigate around the area without a map.
    \item \textit{Intermediate} - Someone with some knowledge about the area might successfully navigate the area.
    \item \textit{Expert} - Someone who could navigate the area without any help.
\end{enumerate}

Using the Suburban route as an example (Fig. \ref{fig:suburb_tbt}), we can compare the total discontinuities and turns between the LLMs and Waze. In this scenario, Waze navigated a distance of 30.2 miles, making a total of 18 turns. The route generated by GPT-4 in Fig. \ref{fig:gpt_suburb} covers a total distance of 39.1 miles, of which 4.2 miles were discontinuous. To reach its destination, GPT-4 made a total of 7 turns. As shown in Figure \ref{fig:gpt_suburb}, most of the discontinuity occurs in the first half of the journey and along the interstates. We believe that some intermediate driving knowledge would be necessary to successfully navigate this route.

For Gemini, the total distance experiencing discontinuity is 28 miles, with a generated distance of 6 miles. Throughout the journey, Gemini made 8 turns to reach its destination. In this case, more than half of the trip consists of discontinuous segments, including a significant discontinuity towards the end, indicating that expert driving skills would be required to navigate the path created by Gemini. 

Mistral experienced a significant discontinuity of 83.3 miles, with only 32.4 miles produced by the model itself. Although Mistral had a total of 11 turns, more than GPT-4 and Gemini, it never reached the intended destination. Successfully completing this route would require expert knowledge.

In multiple cases, there are discontinuities of more than one mile between the planned route's endpoint and the intended destination. In these cases, we indicate that the LLMs did not reach destination: Mistral and Gemini did not reach destination in the suburban TbT case, and none of the LLMs reached destination in the rural TbT case.

\begin{figure*}[]
    \centering
\includegraphics[width=\textwidth]{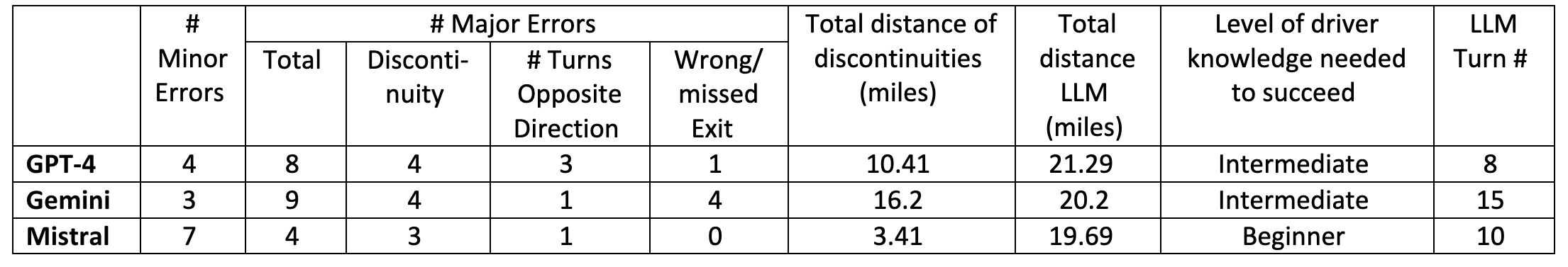}
    \caption{Urban TbT Table (Total distance Waze: 21 miles, Waze turn \#: 12)}
    \label{fig:urban_table}
\end{figure*}

\begin{figure*}[]
    \centering
\includegraphics[width=\textwidth]{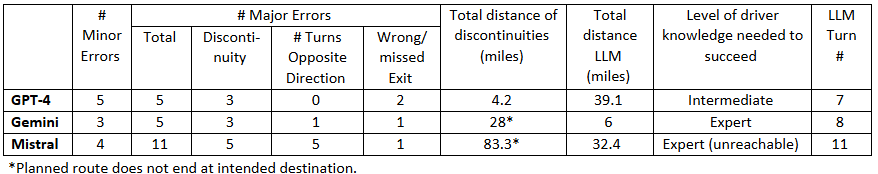}
    \caption{Suburban TbT Table (Total distance Waze: 30.2 miles, Waze turn \#: 18)}
    \label{fig:suburban_table}
\end{figure*}

\begin{figure*}[]
    \centering
\includegraphics[width=\textwidth]{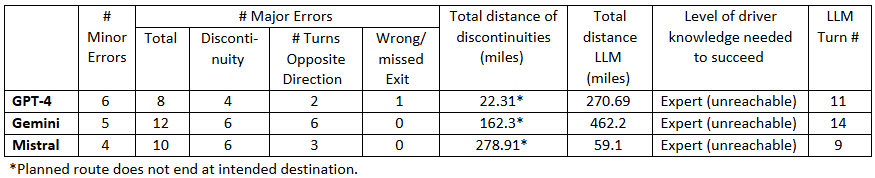}
    \caption{Rural TbT Table (Total distance Waze: 291.4 miles, Waze turn \#: 14)}
    \label{fig:rural_table}
\end{figure*}

\begin{figure*}[]
    \centering    \includegraphics[width=\textwidth]{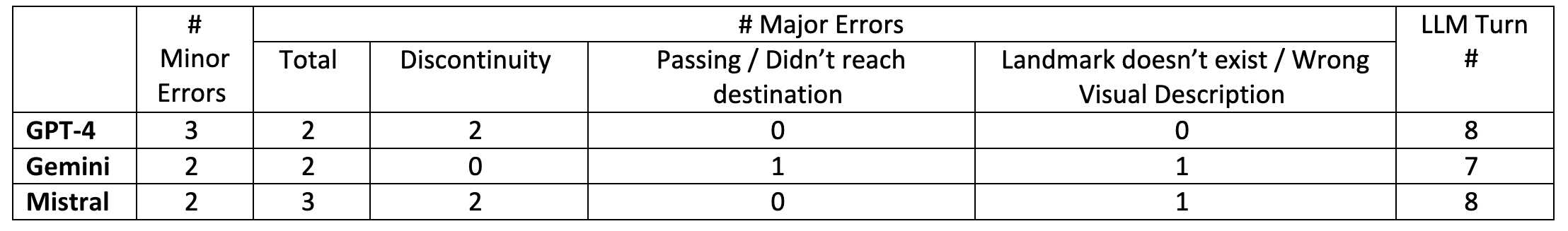}
    \caption{Short VLN Table (Typical student path turn \#: 2)}
    \label{fig:short_vl_table}
\end{figure*}

\begin{figure*}[]
    \centering
\includegraphics[width=\textwidth]{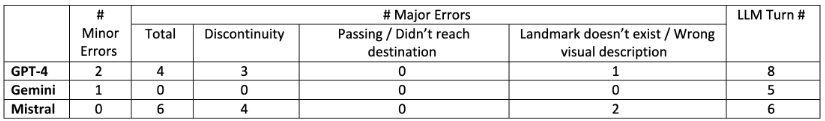}
    \caption{Medium VLN Table (Typical student path turn \#: 11)}
    \label{fig:mid_vl_table}
\end{figure*}

\begin{figure*}[]
    \centering
\includegraphics[width=1\linewidth]{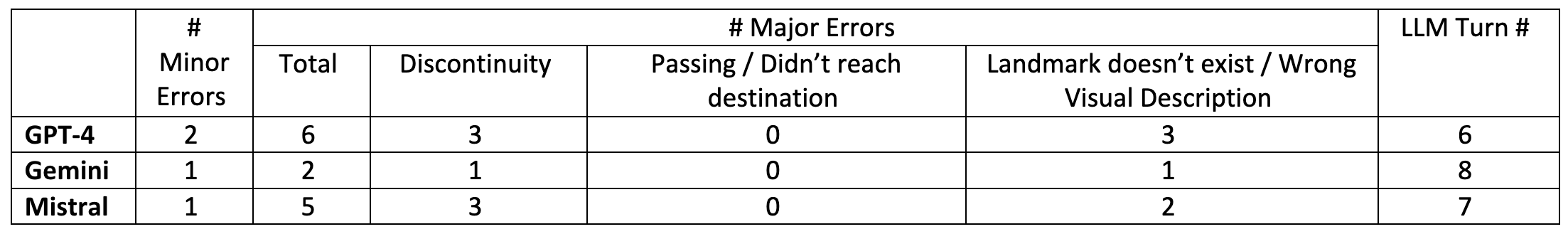}
    \caption{Long VLN Table (Typical student path turn \#: 11)}
    \label{fig:long_vl_table}
\end{figure*}

\subsection{Time constraints}

For the TbT scenarios, we tested whether the LLMs could plan paths with time constraints. If an LLM's response indicated a travel start time and an end time that met the user-specified requirements, then it counted as satisfying the time constraint, regardless of whether the actual planned path contained errors. For example, in the urban TbT scenario, the prompt asked to arrive at the Nationals Stadium two hours before the first weekend game in August 2024. GPT-4 responded, ``The estimated driving time is around 45 minutes to 1 hour, but be sure to account for potential traffic, especially closer to the stadium. For a 4:05 PM game on August 3rd, you should leave by 1:00 PM at the latest." This response satisfied the time constraint.

Among the three tested LLMs, GPT-4 was best at satisfying time constraints. It successfully met time constraints for the urban and suburban scenarios and only failed in the rural scenario. In contrast, Gemini and Mistral did not satisfy any time constraints. Gemini made some attempts but failed. For example, when asked to find the first weekend game, it found the first game on Friday. Mistral, however, could not search the internet in real-time, so it could not find any gaming information.

\subsection{LLM Performance Rankings}
To rank the performance of the LLMs, we collected the total number of errors (Fig. \ref{fig:urban_table}-\ref{fig:long_vl_table}). Given that there are several major error categories, we assigned a weight to each category based on the severity of the errors. For TbT errors, both discontinuities and incorrect directions were assigned weights of 0.4, as these errors are equally severe. Wrong or missed exits were considered less serious and received a weight of 0.2. For VLN errors, if there were discontinuities or if the model failed to reach the destination, weights of 0.8 were assigned. If a landmark did not exist or the description of the landmark was inaccurate, a weight of 0.2 was given.

\begin{table}[h!]
\centering
\begin{tabular}{|p{3.8cm}|p{3.8cm}|}
\hline
\cellcolor{gray!30}\textbf{Urban TbT:} & \cellcolor{gray!30}\textbf{Short VLN:} \\ 
1. Mistral & 1. Gemini \\
2. Gemini & 2. GPT-4 \\
3. GPT-4 & 3. Mistral \\
\hline
\cellcolor{gray!30}\textbf{Suburban TbT:} & \cellcolor{gray!30}\textbf{Medium VLN:} \\ 
1. GPT-4 & 1. Gemini \\
2. Gemini & 2. GPT-4 \\
3. Mistral & 3. Mistral \\
\hline
\cellcolor{gray!30}\textbf{Rural TbT:} & \cellcolor{gray!30}\textbf{Long VLN:} \\ 
1. GPT-4 & 1. Gemini \\
2. Mistral & 2. Mistral \\
3. Gemini & 3. GPT-4 \\
\hline
\end{tabular}
\caption{Performance rankings of the LLMs}
\label{table:llm_ranking}
\end{table}

Table \ref{table:llm_ranking} shows the LLMs' performance rankings, according to the above rules. For TbT scenarios, Mistral excelled in urban navigation, while ChatGPT-4 performed best in suburban and rural environments. Conversely, ChatGPT-4 had the weakest performance in urban navigation, Mistral struggled the most in suburban navigation, and Gemini performed the worst in rural navigation. These results highlight significant limitations regarding the reliability of all models.

In VLN scenarios, Gemini performed the best. In contrast, Mistral struggled with short and medium-distance visual landmark navigation, while ChatGPT-4 had the lowest performance in visual tasking. However, even Gemini made several major and minor errors, suggesting that even the best performer is not reliable as a path-planning tool.

None of the models accurately planned any of the paths and all experienced major errors, demonstrating that they are fundamentally unsuitable for reliable path-planning applications. These findings emphasize the inherent limitations of using LLMs for tasks that require precise navigation and spatial understanding. 

\section{Discussion}

Given that none of the tested LLMs could plan paths in the real world, it begs the question, what could have been done to achieve some success? In this section, we discuss the future research directions. 

\subsection{Reality checks}

One striking result is that some paths were so absurd that even a brief human glance would flag such a wrong solution. For example, in the rural TbT scenario, while Waze's path was 291.4 miles, Gemini planned a path of 462.2 miles. A reality check such as a simple online search about the approximate distance between the destinations may have helped catch such errors. However,  currently LLMs cannot initiate such quality assurance reality checks.

Several researchers have investigated how LLMs could perform fact checks \cite{augenstein2024factuality, zhang2023towards}. However, these studies involve specifically assigning LLMs fact-checking tasks. LLMs are yet to learn when to initiate fact checks or reality checks. This is a critical area of future research for LLMs.

\subsection{In-context transparency}
Our experiments examined LLMs across varying difficulty levels, but the LLMs were not able to adapt to the different levels of complexity and they made more mistakes while attempting the harder tasks. Access to mapping services could help the performance of path-planning LLMs but only Gemini could access the Google Maps API, redirecting users to Google Maps. However, the API could not provide directions in text. So, in this effort, Gemini's paths are not based on Google Maps. Despite this known limitation, however, none of the tested LLMs informed them of this weakness. Instead, they started giving directions right away, and not appropriately warning users of their limitations.

These behaviors could lead users to overtrust the LLMs, which could be very problematic if deployed in actual vehicles. In many cases, overtrust and overreliance on AI systems can lead to disastrous outcomes \cite{kim2024m}. Although currently all the LLMs' user interfaces include a line indicating they may make mistakes (e.g., ``Gemini may display inaccurate info, including about people, so double-check its responses"), the line could easily be ignored by users not paying attention.

Future LLMs should be more transparent about their limitations. In addition to one-line disclaimers, they should indicate their limitations in context. When prompted to complete tasks for which they do not have enough information, they should acknowledge their limitations before attempting the task. They should be able to inform users that they are less certain about such results and abstain from doing such tasks.

Achieving in-context transparency is challenging. Researchers find that LLMs are often overconfident \cite{kim2024m, wen2024human, leng2024taming}, and more research is needed to obtain accurate numerical estimates of uncertainty for LLMs, as well as how well people with various backgrounds understand such estimates.

\subsection{The case for smaller models}

All tested LLMs fell short of path planning, and no LLM consistently outperformed the others. One common argument for improving LLM performance is to increase the input data and associated parameters. However, increasing the model's size does not necessarily lead to better path-planning performance, which raises the question of whether we need to train ever-larger models.

Some researchers have also challenged the idea that larger models equal better models. \citet{hassid2024larger} revealed that repeated use of smaller LLMs could yield consistent improvements for code generation tasks, highlighting the potential of using smaller models instead of larger ones. From an ethical perspective, \citet{bender2021dangers} warned about the risks of training ever-larger models, including environmental costs and reinforcing social stereotypes. 

An alternative path is to train smaller but more fine-tuned models. For example, \citet{aghzal2023can} showed that fine-tuning small models can be an effective approach for some path-planning tasks. However, fine-tuned small models struggle to generalize to more complex environments, suggesting more future research \cite{aghzal2023can}.


\subsection{Limitations}

One limitation of our study is that we only evaluated the LLMs on specific routes in our local area, which may not generalize across broader geographical areas. Furthermore, when new versions of LLMs roll out, new studies will be needed to test their path-planning abilities. In particular, having real-time access to mapping services could be game-changing for the path-planning abilities of future LLMs.

\section{Conclusion}

Although there is much hype around using LLMs for path planning, our experiments show that our tested LLMs are not capable of path planning in the real world. Overall, GPT-4 performed slightly better in TbT scenarios, and Gemini performed better in VLN scenarios, but all LLMs made various mistakes and are far from useful in real-world path planning. Therefore, these LLMs should not be used in directing vehicle navigation, and car companies should take extreme caution in using this experimental software. Going forward, we advocate for more in-context transparency about LLMs' limitations, implementing mechanisms for initiating reality checks, and training smaller and more specialized models.



\bibliography{aaai24}

\begin{thebibliography}{16}
\providecommand{\natexlab}[1]{#1}

\bibitem[{Aghzal, Plaku, and Yao(2023)}]{aghzal2023can}
Aghzal, M.; Plaku, E.; and Yao, Z. 2023.
\newblock Can large language models be good path planners? a benchmark and investigation on spatial-temporal reasoning.
\newblock \emph{arXiv preprint arXiv:2310.03249}.

\bibitem[{Augenstein et~al.(2024)Augenstein, Baldwin, Cha, Chakraborty, Ciampaglia, Corney, DiResta, Ferrara, Hale, Halevy et~al.}]{augenstein2024factuality}
Augenstein, I.; Baldwin, T.; Cha, M.; Chakraborty, T.; Ciampaglia, G.~L.; Corney, D.; DiResta, R.; Ferrara, E.; Hale, S.; Halevy, A.; et~al. 2024.
\newblock Factuality challenges in the era of large language models and opportunities for fact-checking.
\newblock \emph{Nature Machine Intelligence}, 1--12.

\bibitem[{Bender et~al.(2021)Bender, Gebru, McMillan-Major, and Shmitchell}]{bender2021dangers}
Bender, E.~M.; Gebru, T.; McMillan-Major, A.; and Shmitchell, S. 2021.
\newblock On the dangers of stochastic parrots: Can language models be too big?
\newblock In \emph{Proceedings of the 2021 ACM conference on fairness, accountability, and transparency}, 610--623.

\bibitem[{Hassid et~al.(2024)Hassid, Remez, Gehring, Schwartz, and Adi}]{hassid2024larger}
Hassid, M.; Remez, T.; Gehring, J.; Schwartz, R.; and Adi, Y. 2024.
\newblock The Larger the Better? Improved LLM Code-Generation via Budget Reallocation.
\newblock \emph{arXiv preprint arXiv:2404.00725}.

\bibitem[{Hawkins(2024)}]{benz}
Hawkins, A.~J. 2024.
\newblock Mercedes-Benz’s best-in-class voice assistant is getting an AI boost.
\newblock \url{https://www.theverge.com/2024/1/9/24028012/mercedes-benz-mbux-voice-assistant-ai-llm-mbos-ces}.
\newblock Accessed: 2024-11-03.

\bibitem[{Kambhampati(2024)}]{kambhampati2024can}
Kambhampati, S. 2024.
\newblock Can large language models reason and plan?
\newblock \emph{Annals of the New York Academy of Sciences}, 1534(1): 15--18.

\bibitem[{Kim et~al.(2024)Kim, Liao, Vorvoreanu, Ballard, and Vaughan}]{kim2024m}
Kim, S.~S.; Liao, Q.~V.; Vorvoreanu, M.; Ballard, S.; and Vaughan, J.~W. 2024.
\newblock " I'm Not Sure, But...": Examining the Impact of Large Language Models' Uncertainty Expression on User Reliance and Trust.
\newblock In \emph{The 2024 ACM Conference on Fairness, Accountability, and Transparency}, 822--835.

\bibitem[{Latif(2024)}]{latif20243p}
Latif, E. 2024.
\newblock 3P-LLM: Probabilistic Path Planning using Large Language Model for Autonomous Robot Navigation.
\newblock \emph{arXiv preprint arXiv:2403.18778}.

\bibitem[{Leng et~al.(2024)Leng, Huang, Zhu, and Huang}]{leng2024taming}
Leng, J.; Huang, C.; Zhu, B.; and Huang, J. 2024.
\newblock Taming Overconfidence in LLMs: Reward Calibration in RLHF.
\newblock \emph{arXiv preprint arXiv:2410.09724}.

\bibitem[{Meng et~al.(2024)Meng, Wang, Yang, Peng, and Chang}]{meng2024llm}
Meng, S.; Wang, Y.; Yang, C.-F.; Peng, N.; and Chang, K.-W. 2024.
\newblock LLM-A*: Large Language Model Enhanced Incremental Heuristic Search on Path Planning.
\newblock \emph{arXiv preprint arXiv:2407.02511}.

\bibitem[{Schumann et~al.(2024)Schumann, Zhu, Feng, Fu, Riezler, and Wang}]{schumann2024velma}
Schumann, R.; Zhu, W.; Feng, W.; Fu, T.-J.; Riezler, S.; and Wang, W.~Y. 2024.
\newblock Velma: Verbalization embodiment of llm agents for vision and language navigation in street view.
\newblock In \emph{Proceedings of the AAAI Conference on Artificial Intelligence}, volume~38, 18924--18933.

\bibitem[{Valmeekam et~al.(2023)Valmeekam, Sreedharan, Marquez, Olmo, and Kambhampati}]{valmeekam2023planning}
Valmeekam, K.; Sreedharan, S.; Marquez, M.; Olmo, A.; and Kambhampati, S. 2023.
\newblock On the planning abilities of large language models (a critical investigation with a proposed benchmark).
\newblock \emph{arXiv preprint arXiv:2302.06706}.

\bibitem[{Volkswagen(2024)}]{VW}
Volkswagen. 2024.
\newblock World premiere at CES: Volkswagen integrates ChatGPT into its vehicles.
\newblock \url{https://www.volkswagen-newsroom.com/en/press-releases/world-premiere-at-ces-volkswagen-integrates-chatgpt-into-its-vehicles-18048}.
\newblock Accessed: 2024-11-03.

\bibitem[{Wen et~al.(2024)Wen, Xu, Bin, Wolfe, Wang, and Howe}]{wen2024human}
Wen, B.; Xu, C.; Bin, H.; Wolfe, R.; Wang, L.~L.; and Howe, B. 2024.
\newblock From Human to Model Overconfidence: Evaluating Confidence Dynamics in Large Language Models.
\newblock In \emph{NeurIPS 2024 Workshop on Behavioral Machine Learning}.

\bibitem[{Zhang and Gao(2023)}]{zhang2023towards}
Zhang, X.; and Gao, W. 2023.
\newblock Towards llm-based fact verification on news claims with a hierarchical step-by-step prompting method.
\newblock \emph{arXiv preprint arXiv:2310.00305}.

\bibitem[{Zhou, Hong, and Wu(2024)}]{zhou2024navgpt}
Zhou, G.; Hong, Y.; and Wu, Q. 2024.
\newblock Navgpt: Explicit reasoning in vision-and-language navigation with large language models.
\newblock In \emph{Proceedings of the AAAI Conference on Artificial Intelligence}, volume~38, 7641--7649.

\end{thebibliography}

\clearpage
\onecolumn
\appendix

\section*{Appendix A}


\subsection*{GPT-4 Prompts and Responses for Urban TbT Scenario}

\textbf{Prompt:} Plan a trip using turn-by-turn instructions from George Mason University Shenandoah Parking Deck to Garage C of the Nationals Stadium to arrive 2 hours before the first-weekend game in August 2024.

\textbf{Response:}  
To provide a detailed trip plan, I need the exact date and time of the game you plan to attend. If you don't have the exact date yet, please provide an estimated time for planning purposes (e.g., an afternoon or evening game). This will help ensure you arrive 2 hours before the game.

\textbf{Follow-up Q1:} What's the first weekend in August of 2024?

\textbf{Response:}  
The first weekend in August 2024 is August 3rd and 4th. If you are planning to attend a game on either of these dates, please specify the time of the game so I can plan your trip accordingly.

\textbf{Follow-up Q2:} Can you find a game in Nationals stadium on August 3rd or 4th?

\textbf{Response:}  
The Washington Nationals have games scheduled on the first weekend of August 2024 at Nationals Park. They will be playing the Milwaukee Brewers on both August 3rd and 4th. Here are the game details:
\begin{itemize}
    \item Saturday, August 3rd, 2024: The game starts at 4:05 PM.
    \item Sunday, August 4th, 2024: The game starts at 1:35 PM (MLB.com).
\end{itemize}
Given these timings, to arrive 2 hours before the game, you should plan to reach Nationals Park by 2:05 PM on Saturday and by 11:35 AM on Sunday.

\textbf{Turn-by-Turn Directions:}
\begin{itemize}
    \item Start: Shenandoah Parking Deck, George Mason University, Fairfax, VA
    \item Head east on Patriot Cir toward George Mason Blvd.
    \item Turn right onto VA-123 S.
    \item Merge onto I-495 N/Capital Beltway toward Maryland.
    \item Take exit 57B for I-395 N toward Washington.
    \item Continue on I-395 N (passing through parts of DC).
    \item Take exit 6 for C Street SW toward Nationals Park.
    \item Continue on C Street SW.
    \item Turn right onto South Capitol Street SE.
    \item Turn left onto N Street SE.
    \item Turn right onto First Street SE.
    \item Garage C will be on your right, just past Nationals Park.
\end{itemize}

The estimated driving time is around 45 minutes to 1 hour, but be sure to account for potential traffic, especially closer to the stadium. For a 4:05 PM game on August 3rd, you should leave by 1:00 PM at the latest. For the 1:35 PM game on August 4th, you should leave by 10:30 AM.

\subsection*{GPT-4 Prompts and Responses for Easy VLN Scenario}

\textbf{Prompt:} Plan a trip using visual landmarks for people who cannot read road sign instructions from George Mason University Nguyen Engineering Building to Johnson Center.

\textbf{Response:}
\begin{itemize}
    \item \textbf{Starting Point: Nguyen Engineering Building}  
    Begin at the main entrance of the Nguyen Engineering Building. You will see a large, modern building with a distinctive glass facade.

    \item \textbf{First Visual Landmark: The Statue}  
    Walk straight ahead towards the statue of George Mason, which is a prominent bronze statue in a small plaza.

    \item \textbf{Second Visual Landmark: The Clock Tower}  
    Continue walking past the statue and head towards the clock tower, a tall structure with a large clock face, easily visible from a distance.

    \item \textbf{Third Visual Landmark: Mason Pond}  
    From the clock tower, take the path that leads to Mason Pond, a serene body of water with a walking path around it. Follow the path around the pond.

    \item \textbf{Fourth Visual Landmark: The Bridge}  
    After walking around Mason Pond, you will see a small pedestrian bridge. Cross this bridge.

    \item \textbf{Fifth Visual Landmark: Harris Theatre}  
    Once you cross the bridge, you will see the Harris Theatre to your right. It's a building with a brick exterior and a modern design.

    \item \textbf{Sixth Visual Landmark: Student Union Building I (SUB I)}  
    Continue straight ahead until you reach the Student Union Building I (SUB I). This building is noticeable with its large, welcoming entrance and banners.

    \item \textbf{Final Visual Landmark: Johnson Center}  
    From SUB I, walk straight towards the Johnson Center. It's a large, multi-story building with a glass atrium and a prominent entrance.
\end{itemize}

By following these visual landmarks, you can navigate easily from the Nguyen Engineering Building to the Johnson Center without needing to read road signs.

\subsection*{GPT-4 Prompts and Responses for Suburban TbT Scenario}

\textbf{Prompt:} Plan a trip using turn-by-turn instructions from George Mason University Shenandoah Parking Deck to go to the Hogsback Mountain Paintball Center on the next federal holiday when the center is open and in time to spend 4 hours on the range.

\textbf{Response:}  
To plan your trip from George Mason University Shenandoah Parking Deck to Hogsback Mountain Paintball Center on the next federal holiday when the center is open, follow these steps:

\textbf{Next Federal Holiday and Operating Hours}  
The next federal holiday is Independence Day, which falls on Thursday, July 4th. Hogsback Mountain Paintball Center is open every Saturday and Sunday from 10 AM to 5 PM, and for pre-scheduled groups of 8 or more, it is open seven days a week from 10 AM to 5 PM.

\textbf{Travel Time and Schedule}  
To spend 4 hours on the range, you'll want to arrive at 10 AM when the center opens. Given that it takes about 1 hour to drive from George Mason University to Hogsback Mountain Paintball Center, you should plan to leave by 8:45 AM to allow for any unexpected delays.

\textbf{Turn-by-Turn Directions:}
\begin{itemize}
    \item Start at George Mason University Shenandoah Parking Deck.
    \item Head northwest on Patriot Circle toward Mason Pond Drive.
    \item Take the 1st right onto Mason Pond Drive.
    \item Turn left onto University Drive.
    \item Merge onto I-66 W via the ramp to Manassas/Front Royal.
    \item Take exit 40 to merge onto US-15 N/James Madison Hwy toward Leesburg.
    \item Stay on US-15 N for approximately 17 miles.
    \item Turn right onto Hogback Mountain Road (VA-651).
    \item Go two blocks and enter West View Farm. Follow signs to Hogsback Mountain Paintball Center.
\end{itemize}

\textbf{Activity Schedule:}
\begin{itemize}
    \item Arrive by 9:45 AM for setup.
    \item Play from 10 AM to 2 PM.
\end{itemize}

\subsection*{GPT-4 Prompts and Responses for  Medium VLN Scenario}

\textbf{Prompt:} Plan a trip using visual landmarks for people who cannot read road sign instructions from George Mason University Nguyen Engineering Building to INTO Mason Center.

\textbf{Response:} Here's a visual-based guide to help you navigate from the Nguyen Engineering Building at George Mason University to the INTO Mason Center:

\begin{itemize}
    \item \textbf{Starting Point: Nguyen Engineering Building}  
    Exit the Nguyen Engineering Building:
    Locate the main entrance/exit of the Nguyen Engineering Building. It's a large, modern building with a glass facade.

    \item \textbf{Landmark 1: Johnson Center}  
    Head towards the Johnson Center:
    As you exit, you will see a large clock tower and a brick-paved plaza area. Head towards this plaza.
    The Johnson Center (JC) is a prominent building with a distinctive clock tower and a large, open area in front.

    \item \textbf{Landmark 2: George Mason Statue}  
    Walk towards the George Mason Statue:
    From the Johnson Center, look for the statue of George Mason. It’s a large bronze statue situated in a small plaza area with benches.
    Walk towards the statue, which will be straight ahead from the Johnson Center.

    \item \textbf{Landmark 3: Harris Theatre}  
    Proceed to the Harris Theatre:
    From the George Mason Statue, turn to your right. You will see a building with a rounded front entrance – this is the Harris Theatre.
    Walk past the Harris Theatre, keeping it on your right.

    \item \textbf{Landmark 4: Fenwick Library}  
    Continue towards Fenwick Library:
    After passing the Harris Theatre, you will come across the Fenwick Library. It is a tall, modern building with a lot of glass windows.
    Continue walking straight, keeping the Fenwick Library on your left.

    \item \textbf{Landmark 5: Merten Hall}  
    Approach Merten Hall:
    As you walk past Fenwick Library, you will see a large building with a distinctive circular design – this is Merten Hall.
    Walk towards Merten Hall, crossing any pedestrian pathways.

    \item \textbf{Final Destination: INTO Mason Center}  
    Arrive at INTO Mason Center:
    From Merten Hall, look for the INTO Mason Center. It’s located nearby and is marked with clear signage that includes the INTO Mason logo.
    The building is modern with large windows and the INTO Mason logo prominently displayed.
\end{itemize}

\textbf{Summary of Visual Landmarks:}
\begin{itemize}
    \item \textbf{Nguyen Engineering Building:} Glass facade.
    \item \textbf{Johnson Center:} Clock tower and brick-paved plaza.
    \item \textbf{George Mason Statue:} Bronze statue.
    \item \textbf{Harris Theatre:} Rounded front entrance.
    \item \textbf{Fenwick Library:} Tall building with glass windows.
    \item \textbf{Merten Hall:} Circular design.
    \item \textbf{INTO Mason Center:} INTO Mason logo and modern facade.
\end{itemize}

\subsection*{GPT-4 Prompts and Responses for Rural TbT Scenario}
 
\textbf{Prompt:} Plan a trip using turn-by-turn instructions from George Mason University Shenandoah Parking Deck to the Shot Tower State Park to arrive on the second day of guided tours on the week of September 30.

\textbf{Response:}  
To plan your trip from George Mason University Shenandoah Parking Deck to Shot Tower State Park for the second day of guided tours in the week of September 30, follow these turn-by-turn instructions. The second day of guided tours typically falls on a Tuesday, assuming the tours start on a Monday.

\textbf{Departure Point:} George Mason University Shenandoah Parking Deck  
\textbf{Destination:} Shot Tower State Park, 176 Orphanage Dr, Austinville, VA 24312  
\textbf{Estimated Travel Time:} Approximately 4 hours and 30 minutes (subject to traffic and road conditions)

\textbf{Route Instructions:}
\begin{itemize}
    \item Start at George Mason University Shenandoah Parking Deck:
    \item Head west on University Dr toward Patriot Cir.
    \item Turn left onto Patriot Cir.
    \item At the traffic circle, take the 1st exit onto Ox Rd.
    \item Merge onto I-66 W:
    \begin{itemize}
        \item Follow signs for I-66 W/US-50 W/US-29 S.
        \item Merge onto I-66 W via the ramp to Front Royal.
    \end{itemize}
    \item Continue on I-66 W for about 63 miles.
    \item Take exit 1A to merge onto I-81 S toward Roanoke:
    \begin{itemize}
        \item Continue on I-81 S for approximately 185 miles.
    \end{itemize}
    \item Take exit 81 to merge onto I-77 S toward Charlotte:
    \begin{itemize}
        \item Continue on I-77 S for about 8 miles.
    \end{itemize}
    \item Take exit 24 for US-52 S toward Hillsville:
    \begin{itemize}
        \item Merge onto US-52 S/Caron Rd.
        \item Continue to follow US-52 S for about 12 miles.
    \end{itemize}
    \item Turn right onto VA-69 S:
    \begin{itemize}
        \item Follow VA-69 S for about 5 miles.
        \item VA-69 S turns slightly left and becomes VA-638 S.
    \end{itemize}
    \item Arrive at Shot Tower State Park:
    \begin{itemize}
        \item Continue straight onto Orphanage Dr.
        \item The destination will be on your left at 176 Orphanage Dr, Austinville, VA 24312.
    \end{itemize}
\end{itemize}

\textbf{Follow-up Q1:} Can you find the second day of guided tours on the week of September 30 and plan the trip for that day?

\textbf{Response:}  
Guided tours at Shot Tower State Park are offered on Saturdays and Sundays from September 7 through October 27. For the week of September 30, the second day of guided tours will be on Sunday, October 6, 2024.

\textbf{Travel Plan:}  
\textbf{Departure Point:} George Mason University Shenandoah Parking Deck  
\textbf{Destination:} Shot Tower State Park, 176 Orphanage Dr, Austinville, VA 24312  
\textbf{Travel Time:} Approximately 4 hours and 30 minutes

\textbf{Tips for the Trip:}
\begin{itemize}
    \item \textbf{Check Traffic Updates:} Make sure to check for real-time traffic updates before starting your trip.
    \item \textbf{Rest Stops:} Plan for rest stops along the way, especially if driving alone.
    \item \textbf{Guided Tours Schedule:} Confirm the guided tour schedule for Shot Tower State Park in advance to ensure you arrive on time.
    \item \textbf{Travel Essentials:} Carry necessary travel essentials, including snacks, water, and a first aid kit.
\end{itemize}

\subsection*{ChatGPT-4 Prompts and Responses for  Hard VLN Scenario}

\textbf{Prompt:} Plan a trip using visual landmarks for people who cannot read road sign instructions from George Mason University Nguyen Engineering Building to Osher Lifelong Learning Institute (OLLI) at George Mason University.

\textbf{Response:} 

\begin{itemize}
    \item \textbf{Starting Point: Nguyen Engineering Building (Engr)}
    \begin{itemize}
        \item \textbf{Landmark:} The building has a distinctive modern design with large windows and a prominent entrance.
        \item \textbf{Visual Cue:} Look for the statue of a man holding a gear near the entrance.
    \end{itemize}
    
    \item \textbf{Step 1: Walk towards the Johnson Center}
    \begin{itemize}
        \item \textbf{Path:} Exit the Nguyen Engineering Building and head straight towards the central part of the campus.
        \item \textbf{Landmark:} You will pass by the Science and Tech I building on your left.
        \item \textbf{Visual Cue:} Look for the large green area with trees and benches.
    \end{itemize}
    
    \item \textbf{Step 2: Reach the Johnson Center (JC)}
    \begin{itemize}
        \item \textbf{Landmark:} A large building with a clock tower and multiple entrances. It’s a hub for students with a food court, library, and study areas.
        \item \textbf{Visual Cue:} Look for the clock tower and glass façade.
    \end{itemize}
    
    \item \textbf{Step 3: Head towards the Center for the Arts}
    \begin{itemize}
        \item \textbf{Path:} From the Johnson Center, walk straight across the central plaza.
        \item \textbf{Landmark:} You will pass the Fenwick Library on your left.
        \item \textbf{Visual Cue:} The library has a modern glass extension.
    \end{itemize}
    
    \item \textbf{Step 4: Pass the Center for the Arts}
    \begin{itemize}
        \item \textbf{Landmark:} A large building with a distinctive curved front and banners for upcoming performances.
        \item \textbf{Visual Cue:} Look for the banners and the curved architecture.
    \end{itemize}
    
    \item \textbf{Step 5: Walk towards Mason Hall}
    \begin{itemize}
        \item \textbf{Path:} Continue walking past the Center for the Arts.
        \item \textbf{Landmark:} You will see Mason Hall ahead, which has a traditional design with brick and white columns.
        \item \textbf{Visual Cue:} Look for the white columns and brick façade.
    \end{itemize}
    
    \item \textbf{Step 6: Head towards the Finley Building}
    \begin{itemize}
        \item \textbf{Path:} Walk past Mason Hall and continue straight.
        \item \textbf{Landmark:} You will see the Finley Building, which is a smaller building with a simple design.
        \item \textbf{Visual Cue:} Look for the building with a straightforward brick design.
    \end{itemize}
    
    \item \textbf{Step 7: Arrive at Osher Lifelong Learning Institute (OLLI)}
    \begin{itemize}
        \item \textbf{Landmark:} The Osher Lifelong Learning Institute is housed in a smaller building with clear signage outside.
        \item \textbf{Visual Cue:} Look for the OLLI sign and the building with a welcoming entrance.
    \end{itemize}
\end{itemize}

\textbf{Summary of Visual Landmarks:}
\begin{itemize}
    \item \textbf{Nguyen Engineering Building:} Statue of a man holding a gear.
    \item \textbf{Science and Tech I Building:} Large green area with trees and benches.
    \item \textbf{Johnson Center:} Clock tower and glass façade.
    \item \textbf{Fenwick Library:} Modern glass extension.
    \item \textbf{Center for the Arts:} Banners and curved architecture.
    \item \textbf{Mason Hall:} White columns and brick façade.
    \item \textbf{Finley Building:} Simple brick design.
    \item \textbf{OLLI:} Clear OLLI sign and welcoming entrance.
\end{itemize}

This route should help guide those who cannot read road signs by using prominent visual landmarks to navigate through the campus.


\subsection*{Gemini Prompts and Responses for Urban TbT Scenario}

\textbf{Prompt:} Plan a trip using turn-by-turn instructions from George Mason University Shenandoah Parking Deck to Garage C of the Nationals Stadium to arrive 2 hours before the first weekend game in August 2024.

\textbf{Response:}  
Certainly! To plan your trip, I looked for the first weekend game of the Washington Nationals in August 2024. Their first home game in August is on Friday, August 2nd, at 7:15 PM. Here's how to get from George Mason University Shenandoah Parking Deck to Garage C at Nationals Park with an estimated travel time of 1 hour and 15 minutes, arriving 2 hours before the game.

\textbf{Departure Point:} George Mason University Shenandoah Parking Deck  
\textbf{Destination:} Nationals Park Garage C  

\textbf{Turn-by-Turn Directions:}
\begin{itemize}
    \item Depart from George Mason University Shenandoah Parking Deck.
    \item Head west on University Drive for 0.3 miles.
    \item Turn left to stay on University Drive for 0.4 miles.
    \item Turn right to merge onto VA-286 E for 4.9 miles.
    \item Take Exit 15 towards I-95 N/Richmond:
    \begin{itemize}
        \item Keep left at the fork and follow signs for I-95 N/Richmond for 1.3 miles.
        \item Use the left 2 lanes to continue toward I-95 N for 1.7 miles.
    \end{itemize}
    \item Take Exit 54B to merge onto I-395 N for 3.9 miles.
    \item Take Exit 9B to merge onto I-66 E for 2.0 miles.
    \item Take Exit 43B to merge onto I-295 S for 1.2 miles.
    \item Take Exit 1B onto South Capitol St E for 0.8 miles.
    \item Turn right onto Potomac Ave SE for 0.4 miles.
    \item Turn right onto N St SE for 0.2 miles.
    \item Turn left onto Half St SE for 0.4 miles.
    \item Your destination will be on the left. Turn left onto Lots C \& D to arrive at Garage C of Nationals Park.
\end{itemize}

\textbf{Total Distance:} 18.3 miles  
\textbf{Estimated Travel Time:} 1 hour and 15 minutes

\textbf{Follow-up Q1:} What about the plan to arrive 2 hours before the first weekend game in August 2024?

\textbf{Response:}  
To arrive at Nationals Park 2 hours before the first weekend game in August 2024, consider the following:

\begin{itemize}
    \item \textbf{Game Schedule:} The Nationals' first home game in August 2024 is on Friday, August 2nd, at 7:10 PM (MLB Nationals Schedule).
    \item \textbf{Target Arrival Time:} Since the game starts at 7:10 PM, aim to arrive at Nationals Park Garage C by 5:10 PM. This provides a 2-hour buffer to account for traffic and parking.
    \item \textbf{Departure Time:} Use online mapping tools to check live traffic conditions on the day of the game. Based on the route described previously (George Mason University Shenandoah Parking Deck to Nationals Park Garage C via I-495 S), traffic should be manageable with a travel time of around 35 minutes. However, accounting for potential delays, target a departure from George Mason University by 3:00 PM.
\end{itemize}

\textbf{Follow-up Q2:} Friday is not weekend  
\textbf{Response:} Cannot find the first weekend game.

\subsection*{Gemini Prompts and Responses for Suburban TbT Scenario}

\textbf{Prompt:} Plan a trip using visual landmarks for people who cannot read road sign instructions from George Mason University Shenandoah Parking Deck to Hogsback Mountain Paintball Center on the next federal holiday when the center is open and in time to spend 4 hours on the range.

\textbf{Response:}  
Google Maps.

\textbf{Follow-up Q1:} The result should be a written route description using visual landmarks, not Google Maps.

\textbf{Response:}  
Certainly, here's a descriptive route using visual landmarks from George Mason University Shenandoah Parking Deck to Hogsback Mountain Paintball Center:

\textbf{Departure Point:} George Mason University Shenandoah Parking Deck  
\textbf{Destination:} Hogsback Mountain Paintball Center  

\textbf{Landmark-Based Directions:}
\begin{itemize}
    \item Start at the Shenandoah Parking Deck at George Mason University. Look for a red brick building with a large sign that says ``Shenandoah Parking Deck.''
    \item Exit the parking deck and make a left onto University Drive.
    \item Continue on University Drive for about 1.2 miles until you reach a large intersection with traffic lights. Here, you will see a prominent blue water tower on your left.
    \item Turn right at the intersection onto VA-286 E/Fairfax County Parkway. You’ll pass a McDonald's restaurant on the corner.
    \item Follow VA-286 E for about 10.3 miles. You'll pass under railroad tracks and then a large shopping center with a Wegmans grocery store on your right.
    \item Stay straight through the intersection with Braddock Road (VA-620) and continue on VA-286 E.
    \item In about 5.2 miles, you will come to a cloverleaf interchange with Dulles Greenway Toll Road on your left. Stay on VA-286 E.
    \item Continue for 3.7 miles and use the left 2 lanes to merge slowly onto VA-7 W toward Leesburg.
    \item Follow VA-7 W for 2.3 miles. You will pass a gas station and convenience store called Sheetz on your right.
    \item Make a left onto Hogback Mountain Road. There is a green metal sign with white lettering for Hogback Mountain Road.
    \item Follow Hogback Mountain Road for 1.3 miles. The road will become curvy and rural. You will pass a large white house on your right.
    \item Hogsback Mountain Paintball will be on your left. You will see a large wooden sign with a paintball splat design and their logo.
\end{itemize}

\textbf{Follow-up Q2:} How to plan to visit the center on the next federal holiday when the center is open and in time to spend 4 hours on the range.

\textbf{Response:}  
The next federal holiday in the United States is Fourth of July, which falls on a Thursday this year (July 4th, 2024). Since Hogsback Mountain Paintball Center is closed on Thursdays, the ideal day to visit the center would be on the following day, Friday, July 5th.

\begin{itemize}
    \item \textbf{Call the Center to Confirm Hours:} Before heading out, it’s best to confirm the center’s hours of operation for the specific day you plan to visit. You can reach Hogsback Mountain Paintball Center by calling +1 703-777-0057. Their hours are typically from 9:00 AM to 5:00 PM on Fridays.
\end{itemize}

\subsection*{Gemini Prompts and Responses for Rural Tbt Scenario}

\textbf{Prompt:} Plan a trip using turn-by-turn instructions from George Mason University Shenandoah Parking Deck to Shot Tower State Park to arrive on the second day of guided tours during the week of September 30.

\textbf{Response:}  
Google Maps, suggested Tuesday, September 3rd.

\textbf{Follow-up Q1:} The result should be a written route description, not Google Maps. Also, September 3rd is not during the week of September 30.

\textbf{Response:}  
You're right, September 3rd isn't during the week of September 30. Here’s a revised plan with written directions to Shot Tower State Park:

\textbf{Trip Date:} To arrive on the second day of guided tours during the week of September 30, aim for Tuesday, October 1.

\textbf{Estimated Distance and Travel Time:} The drive is approximately 290 miles and takes about 4 hours and 29 minutes.

\textbf{Turn-by-Turn Directions:}
\begin{itemize}
    \item Start at the Shenandoah Parking Deck at George Mason University.
    \item Exit the parking deck and turn right onto University Drive.
    \item Continue straight on University Drive for about 1.5 miles until you reach VA-286 E/Fairfax County Parkway.
    \item Take VA-286 E/Fairfax County Parkway for approximately 12.5 miles.
    \item Merge onto I-66 E via the exit on the left for I-66 East/Washington. Stay on I-66 E for 45 miles.
    \item Take exit 40B for US-50 E/Washington Dulles Toll Rd East. Keep left at the fork and follow signs for US-50 E. This will be a toll road.
    \item Stay on US-50 E for about 134 miles.
    \item Take exit 172 for VA-16 E towards Christiansburg. Stay right at the fork and follow signs for VA-16 E.
    \item Continue on VA-16 E for approximately 37 miles.
    \item Turn left onto Orphanage Dr and continue for about 0.3 miles.
    \item Your destination will be on the right. You have arrived at Shot Tower State Park!
\end{itemize}

\textbf{Follow-up Q2:} September 24 is also not during the week of September 30.

\textbf{Response:} Tuesday, October 1st.

\subsection*{Mistral Prompt and Response for Urban TbT Scenario}

\textbf{Prompt:} Plan a trip using turn-by-turn instructions from George Mason University Shenandoah Parking Deck to Garage C of the Nationals Stadium to arrive 2 hours before the first weekend game in August 2024.

\textbf{Response:}  
Sure, I'd be happy to help you plan your trip. Please note that I'm providing a general plan based on typical traffic patterns and public transportation schedules, but real-time conditions may vary. Also, I don't have real-time access to game schedules, so I'll assume the game is on Saturday, August 3, 2024.

\textbf{Departure Point:} George Mason University Shenandoah Parking Deck, 4400 University Dr, Fairfax, VA 22030  
\textbf{Destination:} Nationals Parking Garage C, 1100 Half St SW, Washington, DC 20024  

\textbf{Turn-by-Turn Directions:}
\begin{itemize}
    \item Start at George Mason University Shenandoah Parking Deck.
    \item Get on I-66 E from University Dr and VA-123 S/Chain Bridge Rd.
    \item Head southeast on Patriot Cir toward Aquia Creek Ln, then turn right onto Aquia Creek Ln.
    \item Turn right onto University Dr and use the left 2 lanes to turn left onto VA-123 S/Chain Bridge Rd.
    \item Use the right lane to take the ramp onto I-66 E.
    \item Follow I-66 E to US-50 E/Constitution Ave NW in Washington.
    \item Merge onto I-66 E and use the right 2 lanes to take exit 72 toward US-50 E/Rosslyn/Memorial Bridge.
    \item Continue onto US-50 E/Arlington Blvd, then onto US-50 E/Constitution Ave NW.
    \item Take 7th St SW to Half St SW by turning right onto 7th St SW, left onto C St SW, then right onto Half St SW.
    \item Arrive at Nationals Parking Garage C.
\end{itemize}

\textbf{Total Distance:} Approximately 25 miles  
\textbf{Estimated Travel Time:} About 40 minutes without traffic. Since it's a weekend trip, allow 2.5 to 3 hours before the game start time to account for potential traffic and parking delays.

\textbf{Additional Note:} Please check the game's start time and verify this route using a real-time navigation service like Google Maps or Waze before you depart. Safe travels!

\textbf{Follow-up Q1:} What's the first weekend in August of 2024?

\textbf{Response:}  
In 2024, the first weekend of August will start on Saturday, August 3rd, and end on Sunday, August 4th. These dates could be useful for planning your trip to Nationals Stadium or any other weekend activities you have planned.

\textbf{Follow-up Q2:} Can you find a game at Nationals Stadium on August 3rd or 4th?

\textbf{Response:}  
I'm an AI and don't have real-time access to sports schedules or the ability to browse the internet. However, you can check the official Major League Baseball (MLB) website or the Washington Nationals' website for their 2024 schedule. Once the schedule is released, you'll be able to see if there's a game on August 3rd or 4th, 2024, and plan your trip accordingly.

\subsection*{Mistral Prompt and Response for Suburban TbT Scenario}

\textbf{Prompt:} Plan a trip using visual landmarks for people who cannot read road sign instructions from George Mason University Shenandoah Parking Deck to Garage C of the Nationals Stadium to arrive 2 hours before the first weekend game in August 2024.

\textbf{Response:}  
Sure, I'd be happy to help you plan this trip. Please note that the exact date of the next federal holiday will depend on the current year, and you should verify that Hogsback Mountain Paintball Center is open on that day. Also, remember to check their operating hours to ensure you arrive in time to spend 4 hours on the range.

Here are the turn-by-turn driving instructions from George Mason University Shenandoah Parking Deck to Hogsback Mountain Paintball Center:
\begin{itemize}
    \item Start at George Mason University Shenandoah Parking Deck, 4400 University Dr, Fairfax, VA 22030.
    \item Head east on Shenandoah Ln toward Patriot Cir.
    \item Turn left onto Patriot Cir.
    \item Turn right onto University Dr.
    \item Use the left 2 lanes to turn left onto Braddock Rd.
    \item Use the right lane to take the Interstate 66 W ramp to Front Royal.
    \item Merge onto I-66 W.
    \item Take exit 43A-43B for US-17 toward Warrenton/Falmouth.
    \item Keep right at the fork, follow signs for US-17 S/Warrenton, and merge onto US-17 S.
    \item Turn right onto VA-29 S/Lee Hwy (signs for Culpeper).
    \item Turn left onto VA-229 S/Rixeyville Rd.
    \item Turn right onto VA-670 W/Viewtown Rd.
    \item Continue onto Hogsback Rd.
\end{itemize}

The destination, Hogsback Mountain Paintball Center, will be on your right: 15540 Hogsback Rd, Amissville, VA 20106.  
This route is approximately 55 miles and takes about 1 hour and 15 minutes without traffic. To ensure you have 4 hours on the range, I recommend checking the paintball center's opening hours and aiming to arrive at least 15-30 minutes before that to allow for any unexpected delays and to get geared up and ready to play.

Please verify all this information before setting out, as road conditions, opening hours, and holiday schedules can change. Safe travels!

\textbf{Follow-up Q1:} Can you find the date of the next federal holiday in 2024 and verify whether Hogsback Mountain Paintball Center is open on that day or not?

\textbf{Response:} I'm an AI and don't have the ability to browse the internet in real-time, so I can't verify the operating hours of Hogsback Mountain Paintball Center on specific dates. However, I can tell you that the next federal holiday in 2024 after the time of this conversation is likely to be Independence Day, which falls on Thursday, July 4, 2024.

To find out if Hogsback Mountain Paintball Center is open on that day, I recommend visiting their official website or contacting them directly via phone or email. This way, you can get the most accurate and up-to-date information about their holiday schedule.

\subsection*{Mistral Prompt and Response for Suburban VLN Scenario}

\textbf{Prompt:} Plan a trip using visual landmarks for people who cannot read road sign instructions from George Mason University Shenandoah Parking Deck to Garage C of the Nationals Stadium to arrive 2 hours before the first weekend game in August 2024.

\textbf{Response:}  
Sure, I'd be happy to help you plan this trip. Please note that the exact date of the next federal holiday will depend on the current year, and you should verify that Hogsback Mountain Paintball Center is open on that day. Also, remember to check their operating hours to ensure you arrive in time to spend 4 hours on the range.

\textbf{Departure Point:} George Mason University Shenandoah Parking Deck, 4400 University Dr, Fairfax, VA 22030  
\textbf{Destination:} Hogsback Mountain Paintball Center, 15540 Hogsback Rd, Amissville, VA 20106  

\textbf{Visual Landmark-Based Directions:}
\begin{itemize}
    \item Begin at George Mason University Shenandoah Parking Deck.
    \item Head east on Shenandoah Ln, passing the distinctive Johnson Center on your right, until you reach Patriot Cir.
    \item Turn left onto Patriot Cir, where you’ll soon see Mason Pond on your right.
    \item Turn right onto University Dr, marked by a large brick archway.
    \item Take the left lane to turn left onto Braddock Rd, then head toward the I-66 ramp.
    \item Look for the tall highway signage marking Interstate 66 W ramp to Front Royal, then merge onto I-66 W.
    \item Continue on I-66 W until you reach a broad green landscape with exit signs, then take exit 43A-43B for US-17 towards Warrenton.
    \item Keep right at the fork, merging onto US-17 S, and look for wide fields to your left.
    \item When you reach a series of stone-lined fences along the road, turn right onto VA-29 S/Lee Hwy.
    \item After a while, turn left onto VA-229 S/Rixeyville Rd, and watch for rolling hills on your right.
    \item Turn right onto VA-670 W/Viewtown Rd, and continue as the road narrows through dense tree-lined sections.
    \item Follow the road as it transitions to Hogsback Rd, where you’ll spot signs for paintball. The paintball center will be on your right.
\end{itemize}

\textbf{Total Distance:} Approximately 55 miles  
\textbf{Estimated Travel Time:} About 1 hour and 15 minutes without traffic. To ensure a full 4 hours on the range, check opening hours and aim to arrive 15-30 minutes early to account for any unexpected delays and to get geared up.

\textbf{Additional Note:} Verify the information about the paintball center’s opening hours and holiday schedule, as these can vary. Safe travels!

\textbf{Follow-up Q1:} Can you find the date of the next federal holiday of 2024 and verify whether Hogsback Mountain Paintball Center is open on that day or not?

\textbf{Response:}  
I'm an AI and don't have the ability to browse the internet in real-time, so I can't verify the operating hours of Hogsback Mountain Paintball Center on specific dates. However, I can tell you that the next federal holiday in 2024 after the time of this conversation is likely to be Independence Day, which falls on Thursday, July 4, 2024.

To find out if Hogsback Mountain Paintball Center is open on that day, I recommend visiting their official website or contacting them directly via phone or email. This way, you can get the most accurate and up-to-date information about their holiday schedule.

\subsection*{Mistral Prompt and Response for Rural TbT Scenario}

\textbf{Prompt:} Plan a trip using turn-by-turn instructions from George Mason University Shenandoah Parking Deck to Shot Tower State Park to arrive on the second day of guided tours on the week of September 30.

\textbf{Response:}  
Sure, I'd be happy to help you plan this trip. Please note that I'm providing a general plan based on typical conditions, but actual traffic and weather conditions can affect your travel time.

\textbf{Departure Date and Time:} The second day of guided tours in the week of September 30 is likely to be October 1, assuming tours start on Monday. To ensure you arrive in time for the tours, I recommend starting your journey early in the morning.

\textbf{Starting Point:} George Mason University Shenandoah Parking Deck, Fairfax, VA  
\textbf{Destination:} Shot Tower State Park, Austin, NV  

\textbf{Initial Directions:}
\begin{itemize}
    \item Head northeast on Shenandoah Ln toward Patriot Cir.
    \item Turn right onto Patriot Cir.
    \item Turn left onto Braddock Rd.
    \item Use the left 2 lanes to turn left onto VA-123 S/Ox Rd.
    \item Use the right lane to take the Interstate 66 W ramp to Front Royal, and merge onto I-66 W.
    \item Continue onto US-50 W/US-340 W/US-17 S/Millwood Ave (follow signs for Winchester).
\end{itemize}

\textbf{Continue on US-50 W across multiple states:}  
This route covers several states (West Virginia, Ohio, Indiana, Illinois, Missouri, Kansas, Colorado, Utah, and Nevada). Take regular rest stops as needed to ensure a safe trip.

\textbf{In Nevada:}
\begin{itemize}
    \item Continue on US-50 W. Take the US-95 N exit toward Reno.
    \item Turn right onto US-95 ALT N.
    \item Turn left onto NV-305 N.
    \item Turn right onto NV-362 E.
    \item Turn left onto NV-722 E.
\end{itemize}

\textbf{Final Destination:} Shot Tower State Park will be on your right.  
This trip spans approximately 2,500 miles and takes around 36 hours without traffic. Given the long distance, I recommend breaking the journey into multiple days, ensuring you rest and stay safe. Please verify the park's opening hours and tour schedules before your trip. Safe travels!

\textbf{Follow-up Q1:} What about the plan to arrive on the second day of guided tours on the week of September 30?

\textbf{Response:} I can't confirm an exact schedule here, so let's assume general availability for now.

\textbf{Follow-up Q2:} Can you check the date and hours of guided tours on the website and tell me how I should schedule my plan?

\textbf{Response:} The park doesn't offer guided tours.

\textbf{Follow-up Q3:} Is Shot Tower State Park in Nevada?

\textbf{Response:} Yes.

\textbf{Follow-up Q4:} Shot Tower State Park is in Virginia.

\textbf{Revised Directions:}
\begin{itemize}
    \item Head northeast on Shenandoah Ln toward Patriot Cir.
    \item Turn right onto Patriot Cir.
    \item Turn left onto Braddock Rd.
    \item Use the left 2 lanes to turn left onto VA-123 S/Ox Rd.
    \item Use the right lane to take the Interstate 66 W ramp to Front Royal, and merge onto I-66 W.
    \item Continue onto US-50 W/US-340 W/US-17 S/Millwood Ave (follow signs for Winchester).
    \item Continue onto US-50 W/US-17 S/John Mosby Hwy.
    \item Turn right onto VA-16 S/VA-42 S (signs for Tye River Turnpike/Amherst).
    \item Turn left onto US-221 S/W Main St.
    \item Turn right onto VA-643 W/VA-652 S/Fort Chiswell Rd.
    \item Turn left onto VA-652 S/Max Meadows Rd.
    \item Turn right onto VA-613 S/Shot Tower Rd.
\end{itemize}

Shot Tower State Park will be on your right.

\end{document}